\documentclass[lettersize,journal]{IEEEtran}
\usepackage{amsmath,amsfonts}
\usepackage{algorithmic}
\usepackage{algorithm}
\usepackage{array}
\usepackage[caption=false,font=footnotesize,labelfont=rm,textfont=rm]{subfig}

\usepackage{textcomp}
\usepackage{stfloats}
\usepackage{url}
\usepackage{verbatim}
\usepackage{graphicx}
\usepackage{cite}
\usepackage{xcolor}  
\usepackage{amsmath}  
\usepackage{booktabs} 
\usepackage{multirow} 
\usepackage{threeparttable}
\usepackage{enumitem}

\begin{document}

\title{A Single-step Accurate Fingerprint Registration Method Based on Local Feature Matching}

\author{Yuwei Jia,~\IEEEmembership{Student Member,~IEEE, }Zhe Cui,~\IEEEmembership{Member,~IEEE, }Fei Su,~\IEEEmembership{Member,~IEEE,}
\thanks{The authors are with the School of Artificial Intelligence, Beijing University of Posts and Telecommunications, Beijing 100876, China, and also with the Beijing Key Laboratory of Network System and Network Culture, Beijing 100876, China.

e-mail: \{jiayw, cuizhe, sufei\}@bupt.edu.cn.
}
}
\markboth{Journal of \LaTeX\ Class Files,~Vol.~14, No.~8, August~2021}%
{Shell \MakeLowercase{\textit{et al.}}: A Sample Article Using IEEEtran.cls for IEEE Journals}


\maketitle

\begin{abstract}
Distortion of the fingerprint images leads to a decline in fingerprint recognition performance, and fingerprint registration can mitigate this distortion issue by accurately aligning two fingerprint images. Currently, fingerprint registration methods often consist of two steps: an initial registration based on minutiae, and a dense registration based on matching points. However, when the quality of fingerprint image is low, the number of detected minutiae is reduced, leading to frequent failures in the initial registration, which ultimately causes the entire fingerprint registration process to fail. In this study, we propose an end-to-end single-step fingerprint registration algorithm that aligns two fingerprints by directly predicting the semi-dense matching points correspondences between two fingerprints. Thus, our method minimizes the risk of minutiae registration failure and also leverages global-local attentions to achieve end-to-end pixel-level alignment between the two fingerprints. Experiment results prove that our method can achieve the state-of-the-art matching performance with only single-step registration, and it can also be used in conjunction with dense registration algorithms for further performance improvements.
\end{abstract}


\section{Introduction}
\IEEEPARstart{T}{he} researches on Automatic Fingerprint Identification Systems (AFIS) have developed over many years, and have made significant contributions in the fields of identity verification and criminal investigation. However, a number of factors still hinder the accuracy of fingerprint recognition algorithms, with fingerprint distortion being one of the critical challenges. When sensors capture fingerprints, the pressure and angle of the fingers on the sensor often cause distortions, leading to significant intra-class variations for fingerprints from the same finger. The goal of fingerprint registration algorithm is to align fingerprints from the same finger as accurately as possible to mitigate the effects of distortion \cite{2009handbook}.

Over the years, fingerprint registration algorithms have evolved from early sparse registration method based on minutiae \cite{tico2003fingerprint} \cite{ross2005deformable} to a more elaborate registration approach that first perform sparse registration followed by dense registration \cite{si2017dense} \cite{cui20182} \cite{cui2019dense} \cite{cui2020dense} \cite{guan2024phase}. These advancements have made it possible to closely align the ridges of two fingerprints. However, most current fingerprint registration algorithms still rely on the sparse registration method based on minutiae as the first step, often referred to as "coarse registration" in some studies. The coarse registration algorithm relies on fingerprint minutiae, which makes it prone to failure 
\begin{figure}[!]
\centering
\centerline{\includegraphics[width=\linewidth]{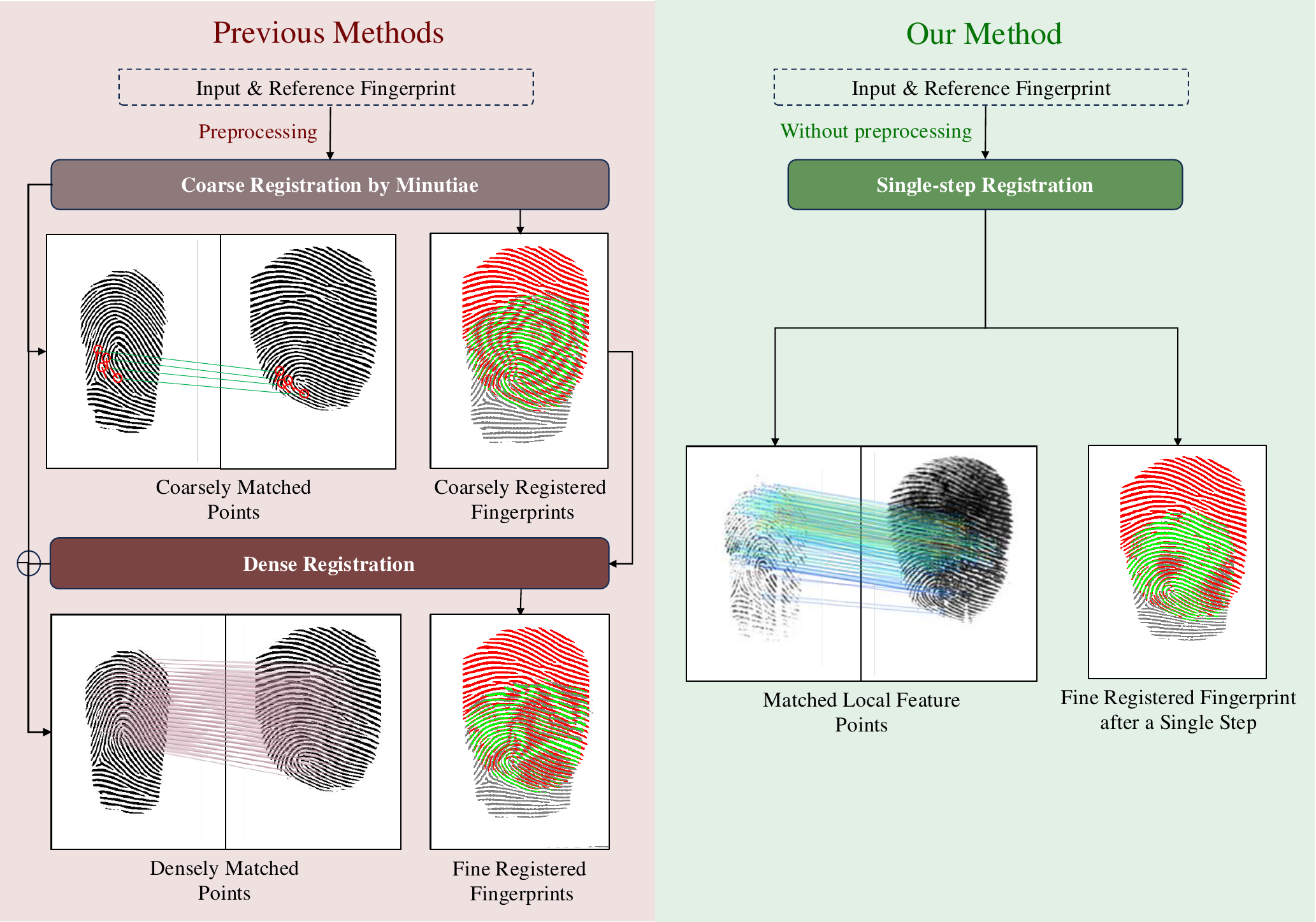}}
\caption{Flowchart of our fingerprint registration method and previous methods. Compared to the previous two-step method, our approach can handle more complex distortions and achieve precise registration in single-step. Left images show the correspondences required for fingerprint registration. In right images, Green areas of indicate overlapping ridges, gray and red indicate non-overlapping ridges of the two fingerprints respectively. }
\label{fig:first_img}
\end{figure} 
when the fingerprint image quality is poor or the minutiae are sparse. Meanwhile, the second step "dense registration" typically focuses solely on improving alignment accuracy, and can only handle small distortion on the basis after coarse registration. If the coarse registration fails, the entire fingerprint registration process is very likely to fail. This limitation significantly reduces the effectiveness of the entire fingerprint registration algorithm for low-quality fingerprints, making fingerprint registration difficult to implement in the real world. In addition, this two-step fingerprint registration approach is computationally demanding, making the entire fingerprint matching process very time consuming.

To address the aforementioned issues, we propose a new accurate and efficient single-step fingerprint registration method, which is a flexible fingerprint registration method based on semi-dense local feature matching \cite{sun2021loftr} \cite{chen2022aspanformer} \cite{wang2024efficient}. Recent studies in the field of fingerprint recognition have demonstrated that semi-dense local feature matching can be effectively used in fingerprint matching \cite{qiu2024ifvit}. However, these studies have not fully leveraged its potential advantages in fingerprint registration. Our method adopts the network structure and training framework from LoFTR\cite{sun2021loftr}, predicting the semi-dense correspondences between two fingerprints. These correspondences are then used to derive a deformation field between the fingerprints. The deformation field is applied to align the input fingerprint with the target fingerprint.

With the help of the Global-Local Attention Block \cite{chen2022aspanformer}, our method is able to align fingerprint pairs as much as possible in a single-step, achieving a higher registration performances and significantly addressing the issue of initial registration failures, thus achieving state-of-the-art matching performance with even reduced runtime. Experimental results demonstrate that the proposed algorithm achieves state-of-the-art matching performance in fingerprint registration.

By using our method as a replacement for the original initial registration module, the total registration performance and matching capability can be significantly improved and outperforms other dense registration methods \cite{guan2024phase}\cite{cui20182}. Compared to other fingerprint registration methods that require operations such as minutiae extraction, phase computation, and binarization, our method allows end-to-end fingerprint registration without preprocessing. By using the same set of deep neural network parameters, our approach can directly register raw fingerprint images across different sensors (optical, thermal wiped, latent), enabling a seamless, cross-sensor registration process without the need for preprocessing steps. 

The main contributions of our work can be summarized as follows:
\begin{itemize}

\item[1)] Semi-dense local feature matching is first applied to the elastic deformation of fingerprints. As an improvement to the coarse registration process in fingerprint registration, this method overcomes the failure issues caused by small overlapping areas and sparse minutiae during coarse registration.

\item[2)] By introducing a Global-Local Attention Block \cite{chen2022aspanformer} into the semi-dense local matching of fingerprints, the accuracy of corresponding point matching is further improved. Thereby for the first time, the matching performance of a single-step fingerprint registration algorithm surpasses that of previous two-step algorithms, significantly reducing the runtime of fingerprint registration.

\item[3)] The proposed fingerprint registration method has good adaptability, capable of registering various types of fingerprints and achieving state-of-the-art performances. Additionally, our algorithm takes raw fingerprint images as input, without the need for preprocessing and feature extraction in previous methods, significantly improving efficiency.

\item[4)] Comprehensive experiments conducted on multiple fingerprint datasets demonstrate that our method can achieve end-to-end registration for fingerprints from different sensors and low-quality samples, breaking the previous bottlenecks in registration accuracy and speed in fingerprint registration research.
\end{itemize}

The paper is organized as follows. Section II reviews the related works. Section III introduces the framework of the proposed fingerprint registration algorithm. Section IV presents the experimental results and discussions. Finally, we make conclusions in Section V.

\section{Related Work}

\subsection{Fingerprint Sparse Registration}

Sparse registration is performed based on a set of control points and a transformation model. By using the correspondences of these control points, the transformation model estimates the deformation field between two images to achieve registration. In the fingerprint domain, these control points are typically the matched minutiae between two fingerprints \cite{bazen2003fingerprint} \cite{ross2005deformable} \cite{almansa2000fingerprint}, and sometimes traditional features like orientation fields \cite{tico2003fingerprint} \cite{chen2009reconstructing}, period map \cite{wan2006fingerprint} or ridge curve \cite{feng2006fingerprint} \cite{choi2011fingerprint} are introduced as additional matching features. 

In earlier approaches, the deformation field estimated by sparse fingerprint registration was rigid, and the transformation algorithm used affine transformations directly. However, researchers later recognized that the surface of the finger skin undergoes elastic deformations, leading to the adoption of Thin Plate Spline (TPS) interpolation to better simulate the elastic deformation of the skin surface. 

In recent years, fingerprint dense registration methods \cite{si2017dense} \cite{cui20182} \cite{guan2024phase} use the same sparse registration algorithms during initial alignment. These algorithms typically estimate the elastic deformation field using Thin-Plate Spline (TPS) when there are many corresponding minutiae, and use direction fields and periodic global search to find the optimal rigid deformation field when there are fewer corresponding minutiae. 
Although this approach reduces the likelihood of initial registration failure to some extent, it is still difficult to achieve a good alignment for two fingerprints with small overlapping areas or low quality.

With the advent of deep neural networks, traditional deformation estimation methods and sparse-registration-based fingerprint coarse registration now have significant room for improvement.

\subsection{Fingerprint Dense Registration}
The motivation behind the invention of fingerprint dense registration algorithms is to improve fingerprint sparse registration algorithms to achieve pixel-level alignment accuracy.

In recent years, significant progress has been made in fingerprint dense registration methods. Si et al. \cite{si2017dense} pioneered the fingerprint dense registration algorithm by performing local correlation matching on image blocks and using Markov Random Fields for global optimization. Cui et al. \cite{cui20182} further improved this algorithm using phase demodulation techniques. Lan et al. \cite{lan2020pre} conducted fingerprint registration based on correlation and direction fields. These methods rely on handcrafted features, which not only struggle to accommodate various forms of test data, leading to performance bottlenecks, but also require several seconds to register a single fingerprint pair.

Cui et al. employed convolutional neural networks (CNNs) for both local \cite{cui2019dense} and global \cite{cui2020dense} fingerprint dense registration, demonstrating the feasibility of deep neural networks in fingerprint dense registration and significantly improving its efficiency. Guan et al.'s PDRNet \cite{guan2024phase}, currently the best-performing fingerprint dense registration algorithm, introduced a unique dual-branch network and incorporated attention mechanisms, raising the performance of fingerprint dense registration to new heights. This method, combining traditional initial registration techniques with Guan et al.'s DDRNet \cite{guan2023regression}, achieves remarkable registration performance, matching accuracy, and efficiency.

Our method is designed to address the limitations of previous initial registration algorithm by combining coarse and dense registrations to a single-step fingerprint registration, allowing it to elevate the overall performance of the fingerprint registration process to a new level.

\begin{figure}[!]
\centering
\centerline{\includegraphics[width=0.75\linewidth]{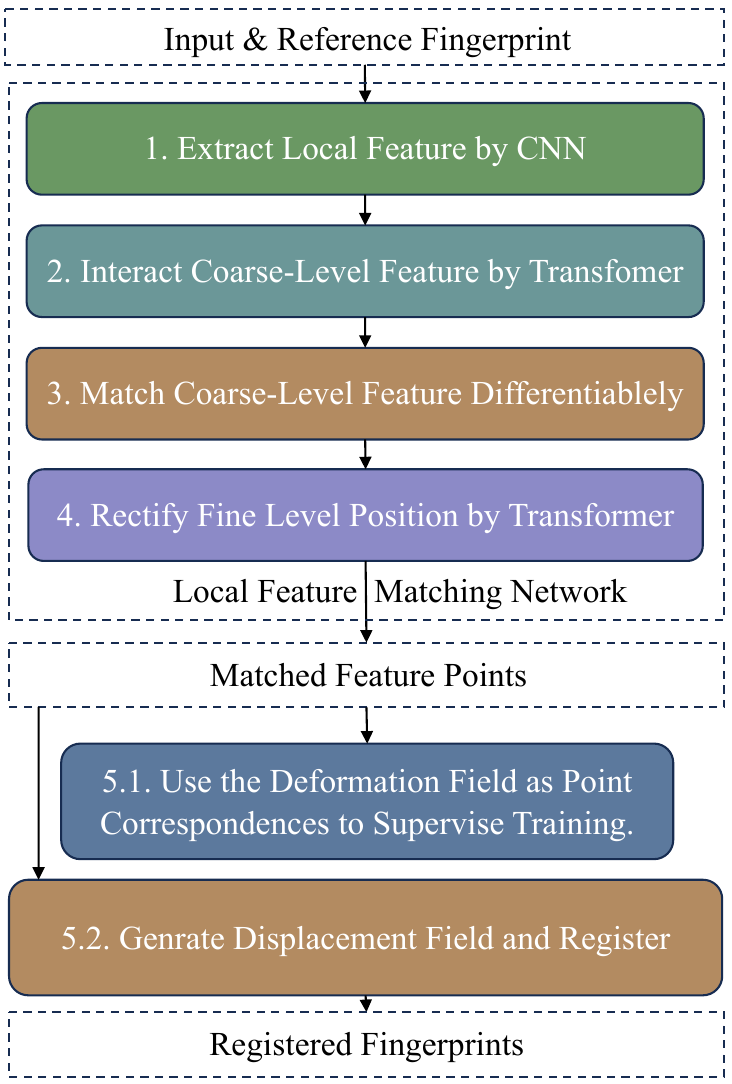}}
\caption{Overview of our proposed fingerprint registration algorithm. We train a Transformer-based network to match local feature points and use those point correspondence to generate deformation field and align fingerprints.}
\label{fig:overview}
\end{figure}

\subsection{Local Feature Matching}
In the field of computer vision, predicting the correspondence of feature points between two images to estimate the affine transformation between them is a common task, referred to as local feature matching. Deep learning-based local feature matching tasks can be categorized into three types: sparse, semi-dense, and dense.

Sparse local feature matching methods\cite{sarlin2020superglue}\cite{lindenberger2023lightglue} are similar to fingerprint minutiae matching. It first detects sparse feature points in an image and then performs binary matching between these feature points. The performance of this method, like in fingerprint registration, is limited by the feature point detection algorithm\cite{detone2018superpoint}. On the other hand, dense local feature matching\cite{edstedt2023dkm}\cite{edstedt2024roma} do not require a feature point detection step and can achieve the best performance. However, its model parameters are large and efficiency is low. 

Semi-dense local feature matching methods\cite{sun2021loftr}\cite{wang2024efficient} strikes a good balance between efficiency and performance. These methods first use twin convolutional neural networks to extract multi-channel features from a pair of images. Then, each pixel from the downsampled feature map is treated as a token and fed into a transformer to predict rough matching relationships. Finally, the rough matching results are passed into higher-resolution feature maps to predict fine-grained matching relationships. This process is analogous to the two-step process in fingerprint registration, but the entire network is trained and inferred end-to-end, which improves efficiency. Aspanformer \cite{chen2022aspanformer} further enhances the accuracy of the matching relationships by introducing a Global-Local Attention module and incorporating optical flow as a local correlation layer.

Recently, Qiu et al. \cite{qiu2024ifvit} introduced LoFTR into the field of fingerprint recognition as an interpretable fingerprint matching method. This approach explored the performance of semi-dense local feature matching in fingerprint matching, but its matching performance was relatively poor, and it did not account for the elastic deformation of fingerprints during registration.

Our method introduces the semi-dense local feature point matching and Global-Local Attention mechanism in fingerprint elastic registration, achieving state-of-the-art fingerprint registration and matching performance. 

\section{Fingerprint Registration Framework}

\begin{figure*}[b]
\centering
\centerline{\includegraphics[width=\linewidth]{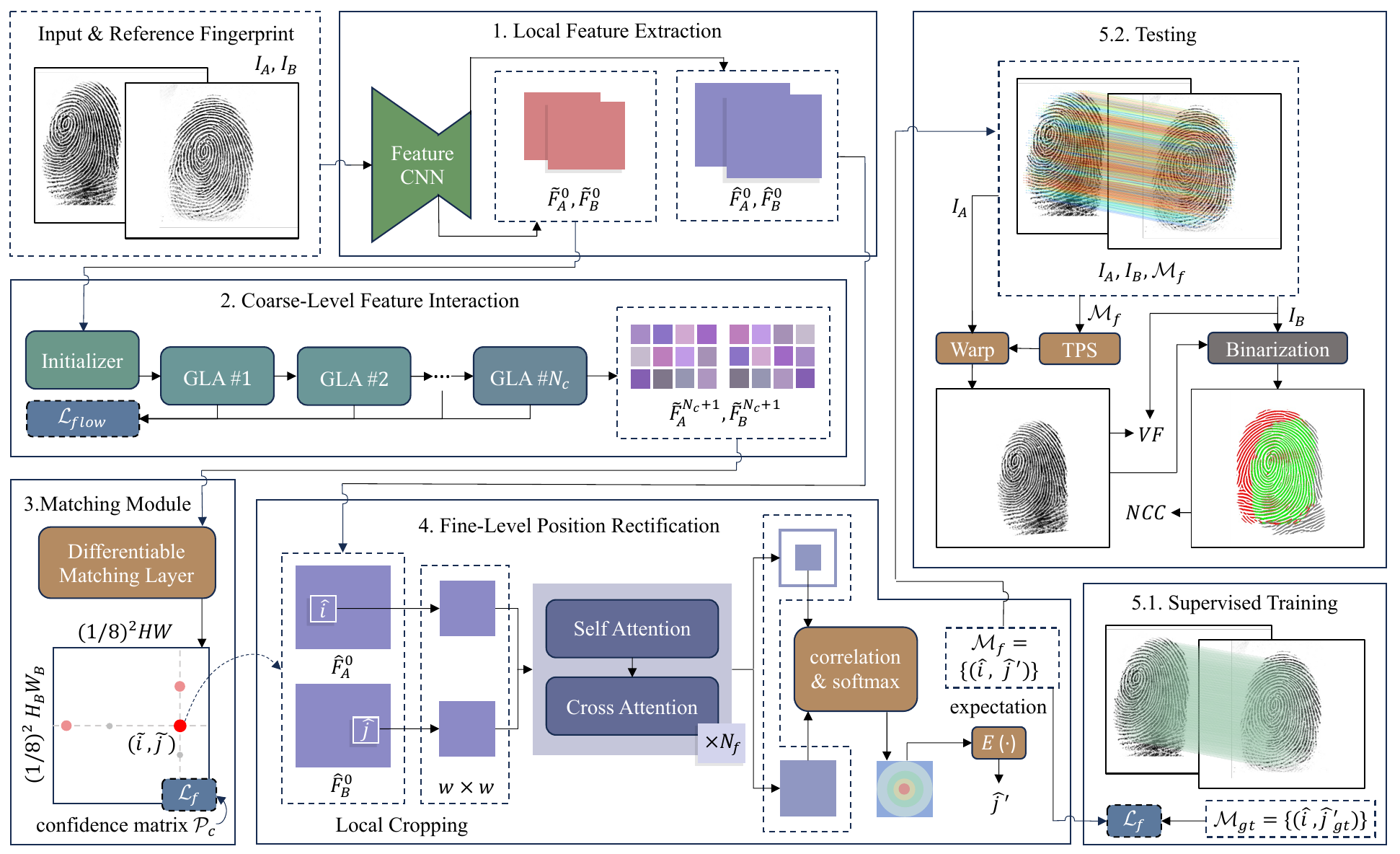}}
\caption{The detailed processes of Fingerprint Semi-dense algorithm}
\label{fig:network}
\end{figure*}

To accurately align the input fingerprint with the reference fingerprint, it is necessary to predict an accurate deformation field between them. To make the estimated deformation filed reliable for fingerprint transformation, Thine-Plate Spline (TPS) is often used to compute the deformation field for smoothing. In our method, we directly predict this deformation field by matching feature points from the two input fingerprints, and the predicted deformation field is used for fingerprint registration.

To construct the matching correspondences, we adopt a Transformer-based local feature matching network. Our work focuses on fingerprint registration rather than fingerprint matching, aiming to align the fingerprint ridges as closely as possible. This imposes higher requirements on the accuracy of the matching points, that previous Transformer-based matching approach \cite{qiu2024ifvit} can not solve. Our method further utilizes the Global-Local Attention module \cite{chen2022aspanformer} which addresses global-local information of attention layers, enabling feature matching network to better capture local details and significantly improving the accuracy of matching points.

Fig. \ref{fig:overview} illustrates the steps of our method. Compared to previous fingerprint registration methods\cite{guan2024phase}\cite{cui20182}\cite{cui2020dense}, which require first performing coarse registration followed by fine registration, our method directly predicts the semi-dense pixel correspondences between a pair of fingerprints using a Transformer-based neural network, and then computes the deformation field from these correspondences. Fig. \ref{fig:network} shows the detailed processes of our algorithm, including the network structure, training, and inference stages. 

\begin{itemize}

\item[] Step 1. We input both the input fingerprints $I_A$ and reference fingerprint $I_B$ into a convolutional neural network to extract the local features. 

\item[] Step 2. They are then passed through a coarse matching network built with Global-Local Attention block for coarse matching of the local features. 

\item[] Step 3. Differentiable bipartite matching is performed based on the coarse matching results. 

\item[] Step 4. The coarse matching results are fed into a transformer for fine matching. 

\item[] Step 5.1. Training: We supervise whether the feature points are correctly matched.

\item[] Step 5.2. Testing: We interpolate the matched points into a TPS for fingerprint registration.
\end{itemize}
The proposed entire registration algorithm is performed end-to-end. 

In this section, we will introduce the detailed modules of the fingerprint semi-dense registration network, including A) Local Feature Extraction, B) Coarse-Level Feature Interaction and Matching, C) Fine-Level Position Rectification, D) Loss Function used in training, and E) Final Registration Step.

\subsection{Local Feature Extraction}

Before feeding the fingerprint to the Transformer for local feature matching, it is necessary to first extract its local features. During Local Feature Extraction, the input fingerprint $I_A$ and reference fingerprint $I_B$ are fed separately into a Feature CNN with an FPN \cite{lin2017feature}. For the fingerprint images at $[H, W]$, the smallest-scale features from the FPN are used as the coarse-level features $\tilde{F}_A^0$ and $\tilde{F}_B^0$ at $[H/8, W/8,C]$, while the final layer features $\hat{F}_A^0$ and $\hat{F}_B^0$ are used as the fine-level features at $[H/2, W/2,C]$. $C$ is the number of feature channels.

\subsection{Coarse-Level Feature Interaction and Matching}
\begin{figure*}[b]
\centering
\centerline{\includegraphics[width=\linewidth]{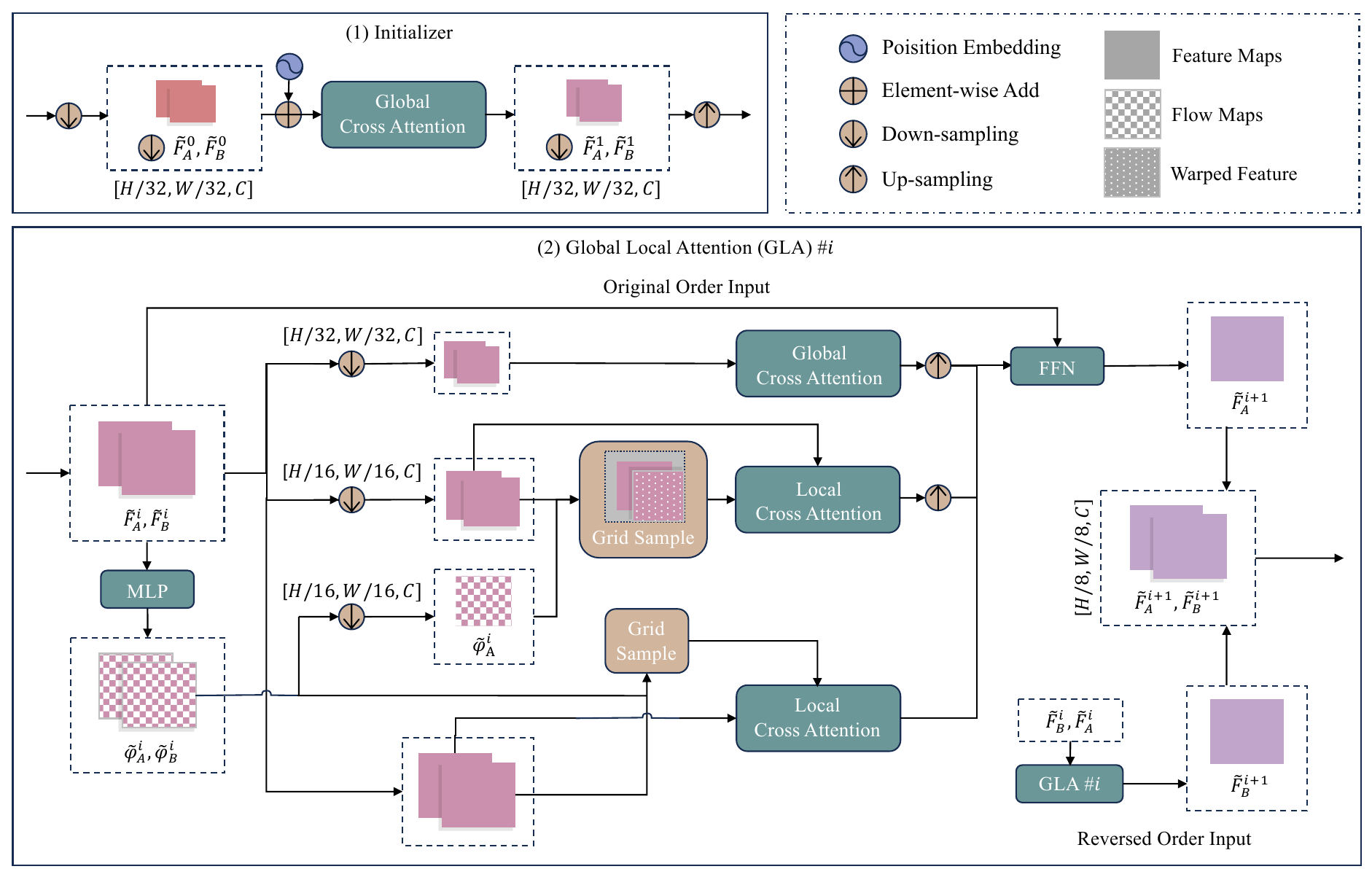}}
\caption{Structure of Global-Local Attention}
\label{fig:gla}
\end{figure*}

The purpose of the coarse matching layer is to match the downsampled local features of the fingerprints. For the coarse matching process, we adopt a transformer structure composed of Global-Local Attention (GLA) modules \cite{chen2022aspanformer}. In this network, the coarse features and their positional encodings are downsampled to $[H/32,W/32]$ for the global attention, and passed into a lightweight cross-attention block, after which they are upsampled back to $[H/8,W/8]$. This component is referred to as the initializer in Fig. \ref{fig:gla}. The initializer features are then fed into a series of GLA modules. 

The final coarse matching features are denoted as $F_A^{N+1}$ and $F_B^{N+1}$ after $N$ GLA blocks. 
Fig. \ref{fig:gla} illustrates the structure of initializer and GLA block in the coarse matching network. The Initializer is mainly responsible for position encoding and initialization of the coarse features. It first downsamples the input coarse features $\tilde{F}_A^0$ and $\tilde{F}_B^0$ to $[H/32,W/32]$, followed by position encoding. The position encoding here transforms positional information into 2D sinusoidal signals of different frequencies and then adds them to the features. Since the input fingerprint images, though varying in size, share a uniform ridge period, there is no need to normalize the position encoding. 

Then, cross-attention is applied to the down-sampled $\tilde{F}_A^0$ and $\tilde{F}_B^0$, followed by upsampling to obtain $\tilde{F}_A^1$ and $\tilde{F}_B^1$.
Subsequently, $\tilde{F}_A^1$ and $\tilde{F}_B^1$ are fed into $N_c$ GLA blocks to obtain the final coarse features, $\tilde{F}_A^{Nc}$ and $\tilde{F}_B^{Nc}$.
For any GLA \#$i$, parts of $\tilde{F}_A^i$ and $\tilde{F}_B^i$ are used to compute the optical flows $\tilde{\varphi}_A^i$ and $\tilde{\varphi}_B^i$, which are then passed through three levels of cross-attention. At the $[H/32, W/32]$ scale, cross-attention is directly applied to $\tilde{F}_A^i$ and $\tilde{F}_B^i$, referred to as Global Cross Attention. 

The results are then upsampled back to $[H/8,W/8]$. At the $[H/16,W/16]$ scale, $\tilde{\varphi}_A^i$ is used to locate the positions of the target features of $\tilde{F}_B^i$ relative to the input features of $\tilde{F}_A^i$. $\tilde{F}_B^i$ is then sampled, aligned with $\tilde{F}_A^i$, and cross-attention is performed between them, referred to as Local Cross Attention, followed by upsampling back to $[H/8,W/8]$. The $[H/8,W/8]$ scale operates similarly to the $[H/16,W/16]$  scale, except no upsampling is needed. 

The features output by the three attention layers are fused using a Feed-Forward Network (FFN) to produce the input features $\tilde{F}_A^{i+1}$ for the next layer. Similarly, the target image features $\tilde{F}_B^{i+1}$ are obtained by swapping the input order of $\tilde{F}_A^i$ and $\tilde{F}_B^i$ for the GLA block. Our positional encoding is the same as that used by Sun et al. \cite{sun2021loftr}.

\textbf{Global Cross Attention}. It's worth noting that the coarse matching network only contains cross-attention, without self-attention. These attention mechanisms can be written as

\begin{equation}
M = \text{Att}(Q,K,V) =\textbf{softmax}(QK^T)V
\end{equation}

where $Q$, $K$, $V$ are linear projections of features from the previous layer, and $M$ is the message output by the attention module. Taking Original Order as an example, for cross-attention, $Q$ comes from the input features $\tilde{F}_A^i$ of the previous layer, while $K$, $V$ come from the target features $\tilde{F}_B^i$ of the previous layer. Subsequently, $M$ and $\tilde{F}_A^i$ are fed into a feed-forward network (FFN) containing concatenation, layer normalization, and linear layers to obtain the input features for the next layer.

\begin{equation}
\tilde{F}_A^{i+1} = \text{FFN}(\tilde{F}_A^{i}, M)
\end{equation}

\textbf{Optical Flow Map}. 
The network structure of LoFTR\cite{sun2021loftr} mainly focuses on global context, with relatively weak attention to local context. However, in fingerprint registration and matching tasks, it is often required to align the ridges of true matching fingerprints as much as possible. Therefore, 
when designing a single-step fingerprint registration network, we pay more attention to the local correspondence between fingerprints.
We try to regress the optical flow in the prediction process of the semi-dense registration network, and then use the optical flow to register the features extracted from fingerprint images during prediction. Applying attention operations to registered features can better enable the network to focus on local areas. 

\begin{figure}[!]
\centering
\centerline{\includegraphics[width=0.75\linewidth]{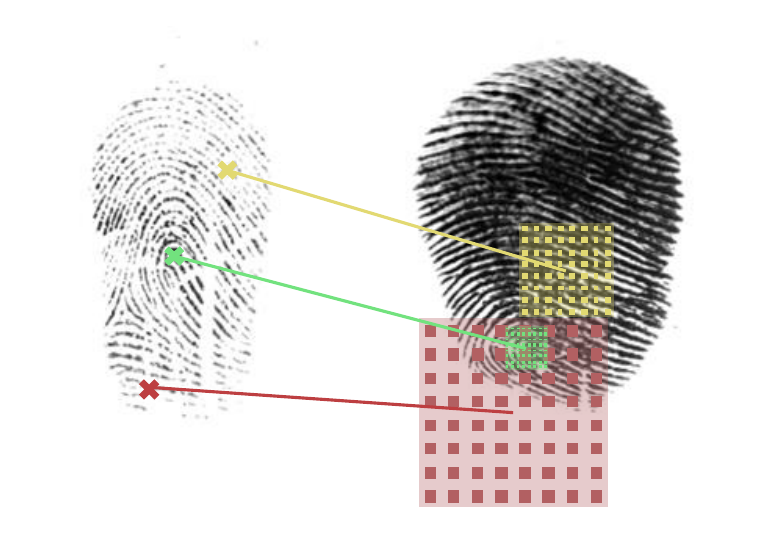}}
\caption{In the Local Cross Attention of GLA, the search region for easily matched points (such as singular points in fingerprints, marked in green in the corresponding image) is small, while the search region for general matching texture areas (marked in yellow in the image) is larger. The search region for points that cannot find corresponding points in the other image (marked in red) is even larger.}
\label{fig:aspan}
\end{figure}

We add a Gaussian distribution-based optical flow prediction module\cite{chen2022aspanformer} to the network. This design can predict different local attention area sizes for local regions with different matching difficulties. For example, in fingerprint registration tasks, regions with higher alignment quality require less context, while regions with larger deformation differences require more context. For background points that don't need alignment, the network will directly search for corresponding points from a larger area, where the similarity of corresponding points is often less than that of regions requiring alignment, as shown in \ref{fig:aspan}.

Specifically, given a pair of images $I_A$, $I_B\in \mathbb{R}^{H\times W}$ corresponding to a pair of optical flows $\varphi_A, \varphi_B \in \mathbb{R}^{H\times W \times 2}$, $I_A[i,j] \leftrightarrow  I_B[\varphi_A[i,j]]$, $I_B[i,j] \leftrightarrow  I_A[\varphi_B[i,j]]$, where $\leftrightarrow$ represents that the pixels in the two images correspond to each other. However, in this section, the optical flow is represented as the mean and standard deviation of a Gaussian distribution, predicting flows $\tilde{\varphi}_A, \tilde{\varphi}_B \in \mathbb{R}^{H\times W \times 4}$, $\tilde{\varphi}[i,j] = [u_{x}^{ij}, u_{y}^{ij},\sigma_x^{ij}, \sigma_y^{ij}]$, where $(u_x^{ij}, u_y^{ij})$ is the mean, also representing the corresponding coordinates of position $(i,j)$ in the other image, and $(\sigma_{x}^{ij}, \sigma_{y}^{ij})$ is the standard deviation, used to determine the size of the attention area. The probability that any point $I_B[x,y]$ in another image corresponds to a point $I_A[i,j]$ in the input image is

\begin{equation}
P(x,y | \varphi_A[i,j])= \frac{1}{2\pi \sigma_{x}^{ij}\sigma_{y}^{ij}} \exp (-\frac{(x-u_x^{ij})^2 }{2{\sigma_x^{ij}}^2 }-\frac{(y-u_y^{ij})^2 }{2{\sigma_y^{ij}}^2 })
\end{equation}

\textbf{Local Cross Attention}. It's easy to understand that the mean of the Gaussian distribution-based optical flow is the position of corresponding points, while the standard deviation will be used for constructing locally adaptive attention areas. Taking the features at $[H/8,W/8]$ scale with origin order $\tilde{F}_A^i$ and $\tilde{F}_B^i$ as an example, we resample $\tilde{F}_B^i$ according to the optical flow, but this sampling process doesn't directly obtain features of the same size as $\tilde{F}_A^i$. Instead, it first divides $\tilde{F}_A^i$ into blocks and then samples from $\tilde{F}_B^i$ for each block. Specifically, each block has a size of $S_1 \times S_1$. Then $\tilde{F}_A^i$ is linearly projected to obtain $Q \in \mathbb{R}^{H\times W\times C}$, from which we can get $H/S_1 \times W/S_1$ blocks, denoted as $Q_p \in \mathbb{R}^{(H/S_1 \times W/S_1)\times S_1^2 \times C}$ . 

Taking any block $Q_{pj} \in \mathbb{R} ^{S_1^2 \times C}$, we can similarly divide $\tilde{\varphi} _A^i$ to get any block $\varphi_{pj} \in \mathbb{R}^{S_1^2 \times 4}$, and obtain $[u_{xj},u_{yj}, \sigma_{xj}, \sigma_{yj}] = \bar{\varphi}_{pj} \in \mathbb{R}^{4}$ by averaging along the first dimension. $\tilde{F}_B^i$ is linearly projected to obtain $K,V \in \mathbb{R}^{H\times W\times C}$. With $(u_{xj}, u_{yj})$ as the center and $(r*\sigma_{xj},r*\sigma_{yj})$ as the width and height, we sample blocks $K_{pj}, V_{pj} \in \mathbb{R}^{S_2^2\times C}$ of size $S_2 \times S_2$, then perform attention operation $m_j = \text{Att}(Q_{pj}, K_{pj}, V_{pj})$, and merge all $m_j$ and reshape to obtain $M$.


\textbf{Matching Module.} The matching layer uses dual-softmax \cite{rocco2018neighbourhood}\cite{tyszkiewicz2020disk}. We flatten $\tilde{F}_A^{N+1}$ and $\tilde{F}_B^{N+1}$ to obtain $\tilde{F}_A \in \mathbb{R}^{N_m\times C}$ and $\tilde{F}_B \in \mathbb{R}^{N_m\times C}$, where $N_m= H/8 \times W/8$. Then we construct the correlation matrix $C = \tau \tilde{F}_A\tilde{F}_B^T$, where $\tau$ is a learnable parameter. We apply dual-softmax of $C$ to obtain the probability for mutual nearest neighbor matching (MNN):
\begin{equation}\mathcal{P}_c = \textbf{softmax}_{row}(C)\cdot \textbf{softmax}_{col}(C)\end{equation}
Then We use MNN criteria to further filter outlier matches, select matches with confidence higher than a threshold of $\theta$ and obtain the coarse matches $\mathcal{M}_c$ :
 \begin{equation}\mathcal{M}_c = \{(\tilde{i},\tilde{j})\vert \forall(\tilde{i},\tilde{j})\in \text{MNN}(\mathcal{P}_c), \mathcal{P}_c(\tilde{i},\tilde{j})\geq \theta  \} \end{equation}

\subsection{Fine-Level Position Rectification}\label{sec:Fine-Level}
Unlike \cite{qiu2024ifvit}, which only focuses on fingerprint matching, our goal is to perform fingerprint registration, which requires precise alignment of the fingerprint ridges. Therefore, we need to obtain more accurate pixel correspondences. We use a correlation-based method for precise registration \cite{yi2016lift}. The purpose of precise matching is to find a more accurate matching point $\hat{j}'$ corresponding to the coarse matching point $\tilde{i}$ from the input fingerprint. The scale of the coarse matching points described earlier is $1/8$  of the original image, so the points must first be converted to the scale of the fine features to obtain $\{\hat{i},\hat{j}\}$. Then, for any $\hat{i}$, we find patches centered around these points with length $w$ in $\hat{F}_A^0$ and $\hat{F}_B^0$ and perform $N_f$ iterations of first self-attention and then cross-attention on the two patches, outputting $\hat{F}_A^{N_f}$ and $\hat{F}_B^{N_f}$. Next, for any $\hat{F}_A^{N_f}$ and $\hat{F}_B^{N_f}$, the center point of $\hat{F}_A^{N_f}$ corresponds to $\hat{i}$ in the original image. The center point of $\hat{F}_A^{N_f}$  is used to compute the similarity for each point in the patch of $\hat{F}_B^{N_f}$, and these similarities are normalized. The expected value is then calculated to find the discrepancy between $\hat{i}$ and $\hat{j}'$, which leads to the computation of $\hat{j}'$. The flatten $\hat{F}_A^{N_f}$ and $\hat{F}_B^{N_f}$ are written as $\hat{F}_A\in \mathbb{R}^{(w\times w)\times C}$ and $\hat{F}_B\in \mathbb{R}^{(w\times w)\times C}$. Let $\hat{J} \in \mathbb{R}^{(w\times w)\times 2}$ represent all the positions of feature points in $\hat{F}_B$,

\begin{equation}\hat{j}' = \textbf{softmax}(\hat{F}_A(\hat{i})\hat{F}_B^T) \hat{J}\end{equation}

The final precise matching result is given by $\{\hat{i},\hat{j}'\}$. Note that here we ignore the difference between the matching result and twice the size of the original image.

\subsection{Loss Function}

Similar to previous works \cite{sarlin2020superglue}\cite{sun2021loftr}\cite{chen2022aspanformer}, our fingerprint registration network's loss function consists of three components:

\begin{equation}
\mathcal{L} = \mathcal{L}_c + \mathcal{L}_f + \alpha \mathcal{L}_{flow}
\end{equation}

where $\mathcal{L}_c$ is the coarse matching loss, $\mathcal{L}_f$ is the fine matching loss, $\mathcal{L}_{flow}$ is the optical flow estimation loss and $\alpha$ is an adjustable parameter.

\begin{equation}
\mathcal{L}_c = -\frac{1}{|\mathcal{M}_{c}^{gt}|} \sum_{(\tilde{i},\tilde{j})\in \mathcal{M}_{c}^{gt}} \log \mathcal{P}_c(\tilde{i},\tilde{j})
\end{equation}

where $\mathcal{M}_{c}^{gt}$ is the ground truth of semi-dense feature point correspondences at scale $[H/8,W/8]$, which can be obtained by quantizing the correspondences constructed in \ref{sec:training_Data_construction}. It's worth noting that $\mathcal{M}_{c}^{gt}$ is an optimal transport matrix. During quantization, points in the target image might be quantized to the same point. To ensure one-to-one correspondence between matching points as much as possible, we deduplicate points in the quantized target image and only keep the corresponding points in the input image. Subsequently, we minimize the negative log-likelihood loss over the grids in $\mathcal{M}_c^{gt}$.

The coordinates of the input image come from a standard grid and won't lose precision due to quantization. However, the coordinates of the target image that are lost due to quantization need further regression. The fine matching loss $\mathcal{L}_f$ is used to correct the imprecise ground truth in coarse matching, which is achieved by calculating the $l_2$ distance between the predicted fine matching position and the ground truth position,

\begin{equation}
\mathcal{L}_f = \frac{1}{\vert \mathcal{M}_f\vert} \sum_{(\hat{i},\hat{j}') \in \mathcal{M}_f} \frac{1}{\sigma^2(\hat{i})} \Vert \hat{j}' -\hat{j}'_{gt} \Vert_2
\end{equation}

where $\mathcal{M}_f$ is the predicted fine matching semi-dense feature point correspondence matrix, $\hat{j}'$ is calculated through the mean of predicted positions, and similarly, we can calculate the variance of predicted positions to obtain $\sigma^2(\hat{i})$. Dividing by $\sigma^2(\hat{i})$ when calculating the $l_2$ distance can help the network better focus on points with lower uncertainty, thereby reducing the network's attention to unmatched points. During training, gradients do not backpropagate from $\sigma^2(\hat{i})$. $\hat{j}'_{gt}$ is the unquantized target position. If $\hat{j}'_{gt}$ is not within a window of size $w$ centered at $\hat{j}'$, it will be ignored when calculating the fine loss.

We also supervise the output flow of each GLA layer in coarse matching, minimizing the negative log-likelihood of its distribution. If the flow of any GLA layer is $\varphi$, the flow loss $\mathcal{L}_{flow}$ for that layer is

\begin{equation}
\mathcal{L}_{flow} = -\frac{1}{\vert D^{gt} \vert } \sum _{ij} \log(P(D_{ij}^{gt} | \varphi_{ij}))
\end{equation}

where $D^{gt}_{ij} = (x_{ij}, y_{ij})$ is the ground truth of the optical flow, which is another representation of $\mathcal{M}_c^{gt}$, and $\varphi_{ij} =(u_x^{ij},u_y^{ij},\sigma_x^{ij},\sigma_y^{ij})$ is the predicted flow at point $(i,j)$. Note that $(\tilde{i},\tilde{j}), (\hat{i},\hat{j}) \in \mathbb{R}^{2\times 2}$ represents the correspondence between coordinates of one image and another, while $(i,j) \in \mathbb{R}^2$ represents the coordinates of any point in one image. If we write $\mathcal{L}_{flow}$ in the form of a Gaussian distribution, we get
\begin{equation}
\begin{aligned}\mathcal{L}_{flow} &= - \frac{1}{ \vert D^{gt} \vert } \sum_{ij} \log [\frac{1}{2\pi \sigma^{ij}_x\sigma_{y}^{ij}}\exp( \\ &-\frac{(x_{ij}-u_{x}^{ij})^2}{2{\sigma_x^{ij}}^2} -\frac{(y_{ij}-u_{y}^{ij})^2}{2{\sigma_y^{ij}}^2})]  \\ &=  \frac{1}{ \vert D^{gt} \vert } \sum_{ij} [\log2\pi + \log{\sigma_{x}^{ij}} + \log{\sigma_{y}^{ij}}  \\&+\frac{(x_{ij}-u_{x}^{ij})^2}{2{\sigma_x^{ij}}^2} +\frac{(y_{ij}-u_{y}^{ij})^2}{2{\sigma_y^{ij}}^2}]\end{aligned}
\end{equation}
As can be seen, during training, this loss not only minimizes the $l_2$ distance between the predicted flow and ground truth flow but also reduces the variance of flow prediction. This results in smaller variances between matching points that are easy to predict, and larger variances between matching points that are more difficult to predict. Points that cannot be matched will get flow prediction results with very large variances, thus achieving adaptive local cross-attention.

\subsection{Final Registration Registration}

During training, we only match semi-dense feature points. To align two fingerprint images, we need to generate a deformation field based on the matching points. We use Thin-plate spline (TPS) to directly interpolate matching points to obtain the deformation field. Compared to previous fingerprint registration methods that directly minimize bending energy using TPS, we add a regularization term to control the distance between warped input points and target points to prevent ridge breaks. Thus, our TPS aims to minimize

\begin{equation}
E[f] = \sum_{i=1}^m  \vert f(\boldsymbol{x}_i) - y_i \vert + \lambda \int_{\mathbb{R^n}} \vert D^2f\vert ^2 dX
\end{equation}

In implementation, this is equivalent to adding a diagonal matrix $\lambda \boldsymbol{I}$ to the distance matrix during TPS solving.

During inference, we use the original images to predict semi-dense local feature matching relationships and perform fingerprint registration.

\section{EXPERIMENTS}

In this section, we conduct comparative experiments with the current state-of-the-art fingerprint registration methods: TPS transformation based on matching minutiae, traditional phase-based registration\cite{cui20182} method, PDRNet\cite{guan2024phase} as a representative of deep neural network-based method, and IFVit\cite{qiu2024ifvit}, a recent Transformer-based fingerprint matching method. Comparisons are performed from three perspectives: matching performance, registration performance, and time efficiency. Ablation experiments are conducted to provide further evidence of the effectiveness of our design.

\subsection{Datasets Description}\label{sec:training_Data_construction}

We test our method on multiple fingerprint datasets, including FVC2004 DB1\_A, FVC2004 DB2\_A, FVC2004 DB3\_A\cite{maio2004fvc2004}, and NIST SD 27\cite{garris2000nist}. Table \ref{table:datasets} provides detailed information about the datasets we used. For training dataset, we select a portion of the data from NIST SD 302 \cite{fiumara2019nist}. We will explain in detail how we conduct the testing and training in the following paragraphs.

\begin{table*}[!]

\caption{Datasets used in this paper}
\label{table:datasets}
\centering
\begin{tabular}{lcccccc}
\toprule
Dataset       & \multicolumn{3}{c}{NIST SD 302}                                                                                                                                                          & \multicolumn{3}{c}{FVC2004 DB1\_A}                                                                                        \\ \midrule
Image          &     \includegraphics[height=2cm]{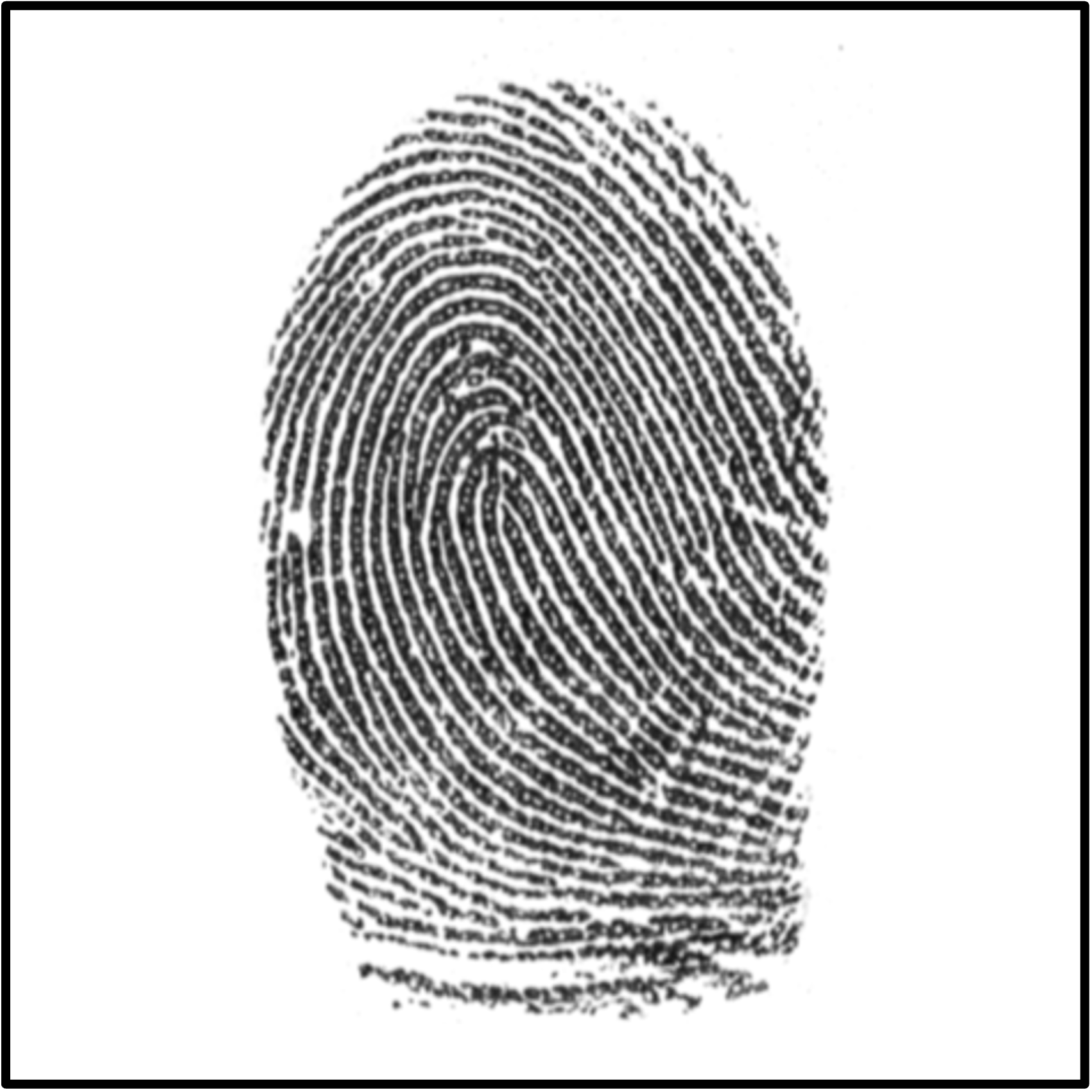}    &             \includegraphics[height=2cm]{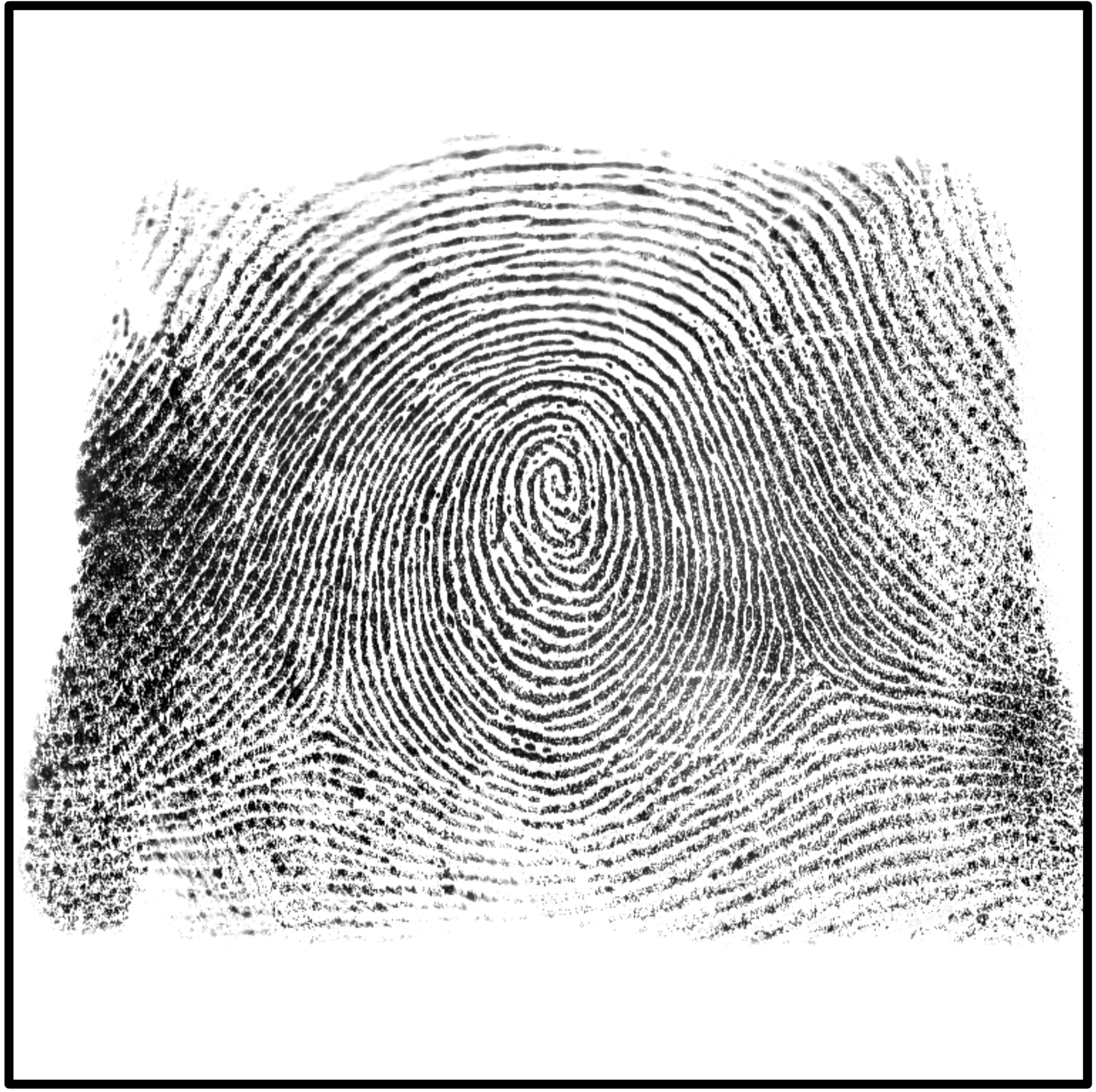}                                                             &                     \includegraphics[height=2cm]{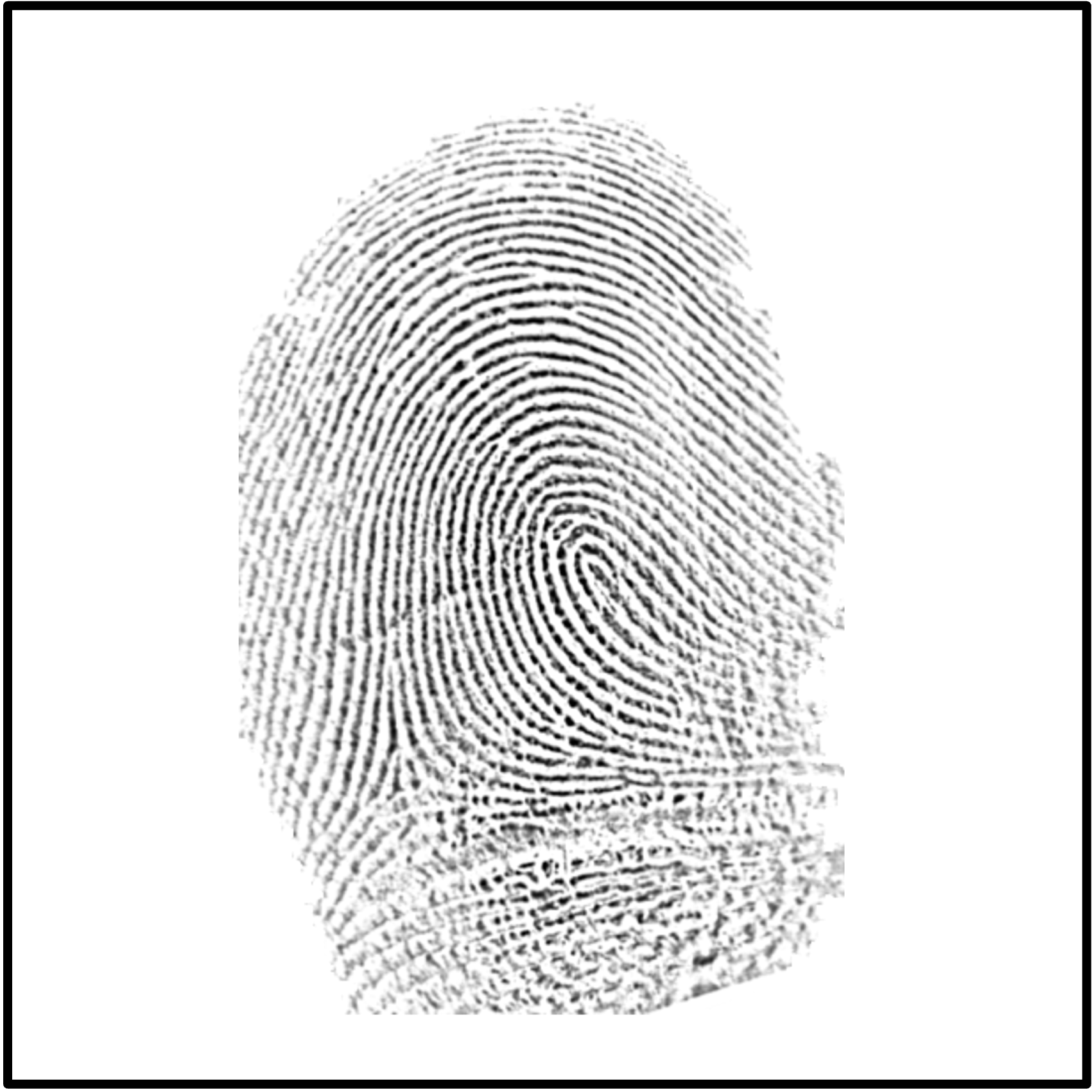}                                           &  \includegraphics[height=2cm]{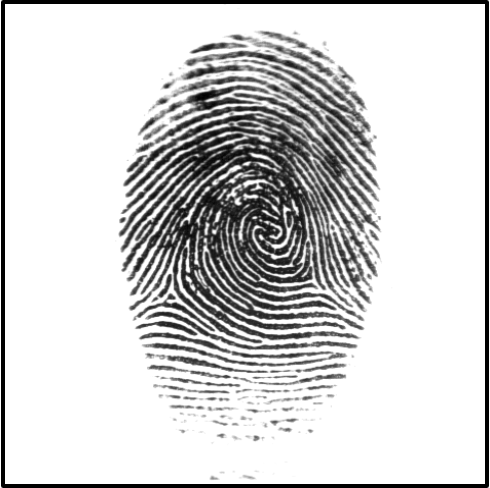}                                      &     \includegraphics[height=2cm]{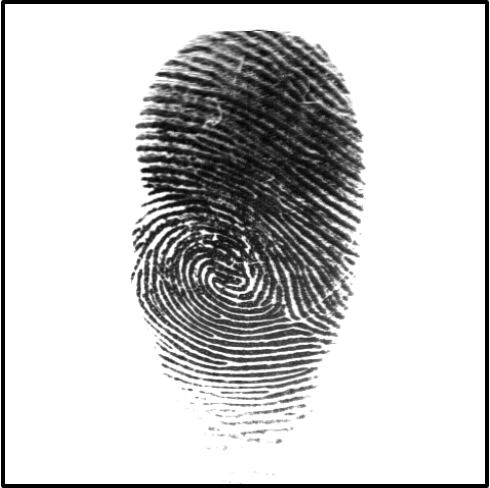}                                     &       \includegraphics[height=2cm]{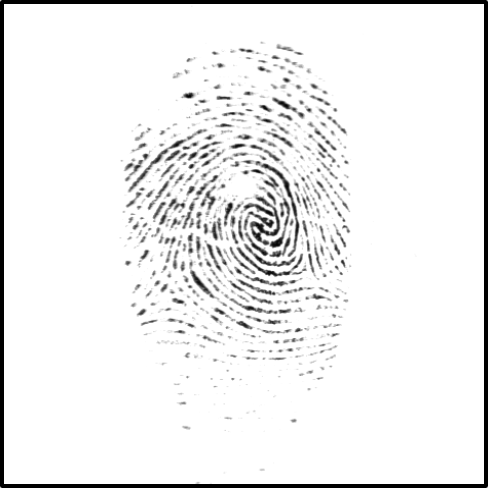}                                   \\ \midrule
Type           & \multicolumn{3}{c}{Roll / Plain}                                                                                                                                                         & \multicolumn{3}{c}{Plain}                                                                                                 \\ \midrule
Sensor         & Optical                                                   & Solid State                                                   & Touch Free                                                 & \multicolumn{3}{c}{Optical}                                                                                               \\ \midrule
Description    & \multicolumn{3}{c}{\begin{tabular}[c]{@{}l@{}}A larger-scale open-source dataset containing fingerprint images \\ collected from more than ten different types of sensors.\end{tabular}} & \multicolumn{3}{c}{\begin{tabular}[c]{@{}l@{}}100 fingers $\times$ 8\\ large distortion and various quality\end{tabular}} \\ \midrule
Usage          & \multicolumn{3}{c}{Training}                                                                                                                                                             & \multicolumn{3}{c}{\begin{tabular}[c]{@{}l@{}}Registration Experiment\\ Matching Experiment\end{tabular}}                  \\ \midrule
Genuine Match  & \multicolumn{3}{c}{$39,476^a$}                                                                                                                                                           & \multicolumn{3}{c}{$2,800^b$}                                                                                             \\ \midrule
Imposter Match & \multicolumn{3}{c}{\textbackslash{}}                                                                                                                                                     & \multicolumn{3}{c}{$4,950^c$}                                                                                             \\ \midrule
\end{tabular}

\begin{tabular}{lcccccc}
\toprule
Dataset       & \multicolumn{3}{c}{FVC2004 DB2\_A}                                                                                        & FVC2004 DB3\_A                                                                       & \multicolumn{2}{c}{NIST SD 27}                                                                                     \\ \midrule
Image          &      \includegraphics[height=2cm]{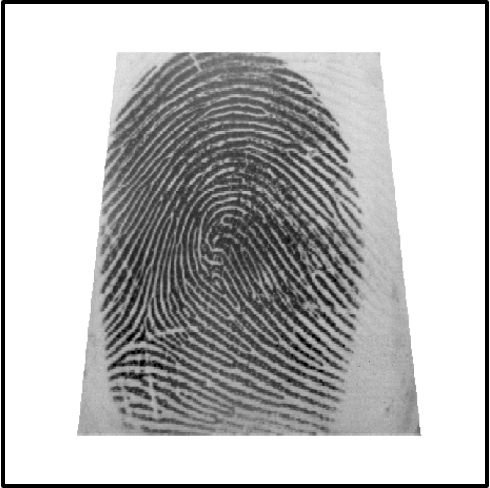}                                   &        \includegraphics[height=2cm]{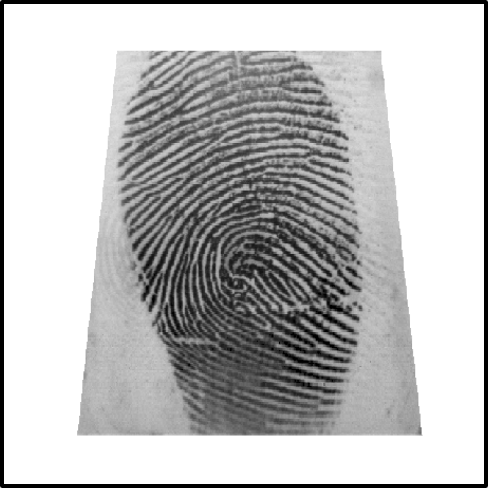}                                &    \includegraphics[height=2cm]{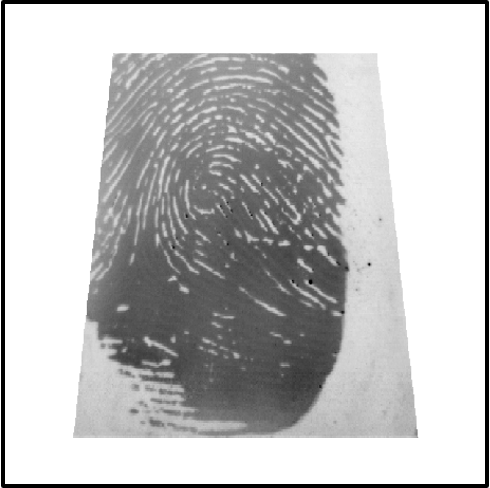}                                      &     \includegraphics[height=2cm]{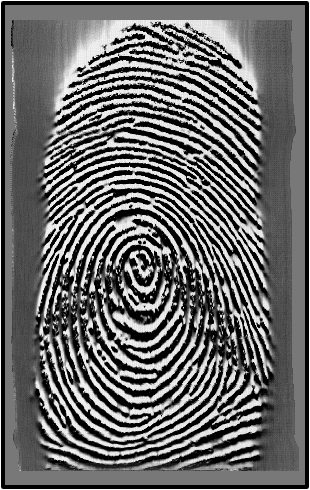} \quad\includegraphics[height=2cm]{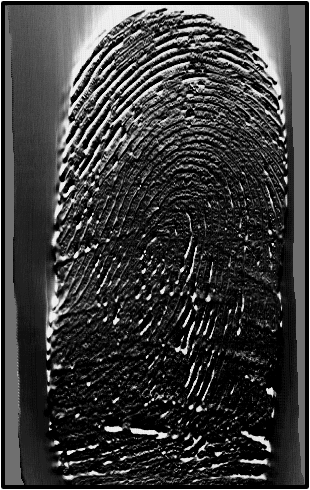}                                                                                  &   \includegraphics[height=2cm]{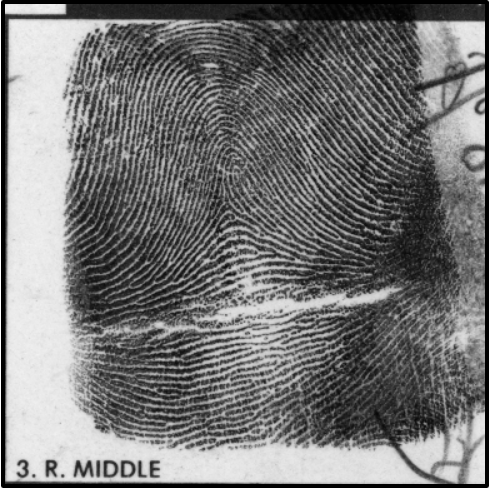}                                                          &     \includegraphics[height=2cm]{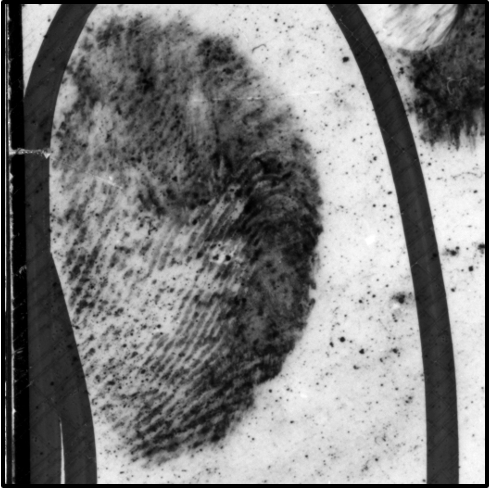}                                                    \\ \midrule
Type           & \multicolumn{3}{c}{Plain}                                                                                                 & Plain                                                                                & Rolled                                                     & Latent                                                \\ \midrule
Sensor         & \multicolumn{3}{c}{Optical}                                                                                               & Thermal sweeping                                                                     & Solid State                                                & -                                                     \\ \midrule
Description    & \multicolumn{3}{c}{\begin{tabular}[c]{@{}c@{}}100 fingers $\times$ 8\\ large distortion and various quality\end{tabular}} & \begin{tabular}[c]{@{}c@{}}100 fingers $\times$ 8\\ various quality\end{tabular}    & \multicolumn{2}{l}{\begin{tabular}[c]{@{}c@{}}258 pairs\\ latent fingerprints from real crime scenes\end{tabular}} \\ \midrule
Usage          & \multicolumn{3}{c}{\begin{tabular}[c]{@{}c@{}}Registration Experiment\\ Matching Experiment\end{tabular}}                  & \begin{tabular}[c]{@{}c@{}}Registration Experiment\\ Matching Experiment\end{tabular} & \multicolumn{2}{c}{Matching Experiment}                                                                           \\ \midrule
Genuine Match  & \multicolumn{3}{c}{$2,800^b$}                                                                                             & $2,800^b$                                                                            & \multicolumn{2}{c}{$258^b$}                                                                                        \\ \midrule
Imposter Match & \multicolumn{3}{c}{$4,950^c$}                                                                                             & $4,950^c$                                                                            & \multicolumn{2}{c}{$66,306^d$}                                                                                     \\ \midrule
\end{tabular}
\begin{tablenotes}
\item $^a$: We selected fingerprint pairs from NIST SD 302 with a two-step registration process and a correlation coefficient greater than 0.7, ultimately retaining 39,476 pairs.
\item $^b$: Match genuine fingerprint pairs pairwise, without considering the exchange of the matching order.
\item $^c$: Each finger's first image is matched with each other, without considering the exchange of the matching order.
\item $^d$: Each latent fingerprint is matched with all rolled fingerprints from other fingers.
\end{tablenotes}

\end{table*}

Previous datasets for dense fingerprint registration were created by applying elastic deformation fields to a single fingerprint image to generate a pair of fingerprint for training. However, such methods failed to enable end-to-end training of fingerprint registration networks due to the two key issues. On one hand, due to the limitations of the training set and the high accuracy requirements for the deformation field, dense registration methods can only handle small deformations, thus struggling to handle large-scale deformations. On the other hand, previous dense registration methods were trained on binarized fingerprints, which made it difficult to adapt to input images of varying quality.

To address these issues, we construct a new fingerprint registration training set based on genuine fingerprint pairs from the public fingerprint dataset NIST SD 302.
Our construction process is done using the previous state-of-the-art two-step fingerprint registration method. We first padd all fingerprints in NIST SD 302 to a size of 800×800. Then, we apply the PDRNet\cite{guan2024phase} registration process to all genuine fingerprint pairs. During this process, fingerprint pairs with registration correlations below 0.7 are discarded. 
When constructing the ground truth for semi-dense feature matching, the input fingerprints $I_A$ and $I_B$ are inputted into a two-step registration, and get a coarse deformation field $D_c$ and a fine deformation field $D_f$. Then we combine the deformation fields from the two stages into a single deformation field. The combination method is defined as:
\begin{equation}
\label{first_merging}
D(x,y) = D_c(x+D_f(x,y)_x, y+D_f(x,y)_y) +D_f(x,y)
\end{equation}, $D_f(x,y)_x$ denotes the deformation of $D_f$ along the x-axis at the point $(x,y)$, while $D_f(x,y)_y$ represents the deformation of  $D_f$ along the y-axis at the same point $(x,y)$.

\begin{figure*}[!]
\centering
\centerline{\includegraphics[width=0.7\linewidth]{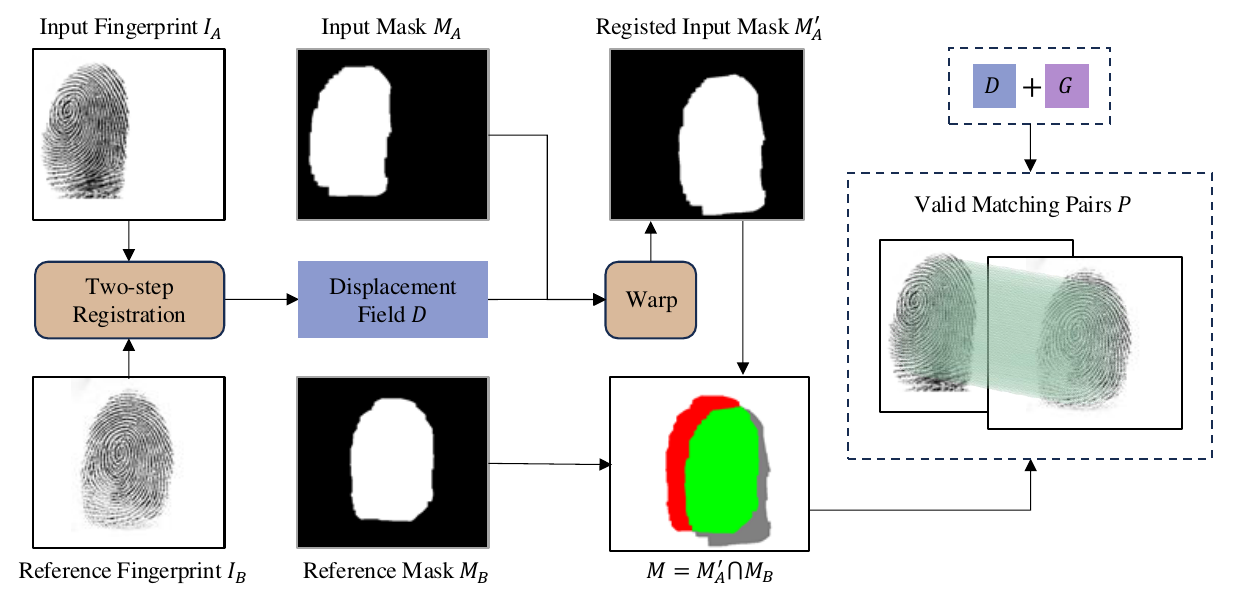}}
\caption{The process of generating ground truth for supervising the training. Here, $D$
  represents the result of merging the deformation fields from the two-step registration, and 
$G$ represents sampling points by the standard grid.}
\label{fig:get_gt}
\end{figure*}

As shown in Fig. \ref{fig:get_gt}, to obtain the ground truth for training, we need to establish the point correspondence $P$ between the overlapping regions of the two fingerprints. This correspondence can be derived by summing the deformation field $D$ and the sampling points by the standard grid $G$. 

However, since $D$ as a deformation field includes both the deformations of the fingerprint texture region and the background region, the corresponding points in the background need to be removed. This can be achieved by applying the deformation field to the masks of the input fingerprints. Let the masks corresponding to $I_A$ and $I_B$ be $M_A$ and $M_B$, respectively. By applying the two-step registration deformation field to $M_A$, we obtain the transformed mask $M_A'$. Next, the intersection region $M = M_A\And M_B $ is computed. The ground truth correspondences for training, $P$
 , is then obtained by retaining only the region of $D+G$ within $M$. This whole process is shown in Fig. \ref{fig:get_gt}

Considering the exclusion of fingerprint pairs with insufficiently precise registration, our training set primarily contains pairs with certain deformation and noise differences. To simulate more complex scenarios, data augmentation is employed during training. The augmentations include random rigid transformation, random exchange and random occlusion.

\subsection{Evaluation Protocols}

Similarly to previous methods \cite{guan2024phase}\cite{cui20182}\cite{cui2020dense}, we use image correlator and Verifinger as evaluation metrics for the similarity score between the two registered fingerprints. For input fingerprint $I_A$ and target fingerprint $I_B$, we first extract the intersection $M$ of two fingerprint image masks $M_A$ and $M_B$. The correlation coefficient is calculated as:

$$
NCC = \frac{\sum_M(I_A-\overline{I_A})\cdot (I_B - \overline{I_B})}{\sqrt{\sum_M (I_A-\overline{I_A})^2\cdot \sum_M(I_B-\overline{I_B})^2}}
$$

where $\overline{I}$ represents the mean value of the parts of an image belonging to $M$. We calculate the $NCC$ between two images as the Image correlator. Meanwhile, we use Verifinger 13.0 \cite{verifinger} to calculate the VeriFinger matching score. The obtained image correlation scores and Verifinger scores are used to calculate the matching performances (FMR and EER).
\subsection{Matching Accuracy}

A main advantage of our algorithm compared to previous dense registration methods is the improvement in matching accuracy. This is because our method advances fingerprint initial registration by direct fingerprint feature matching, solving many cases that previous initial minutiae registration algorithms could not handle. As shown in Table \ref{table:corr_match} and \ref{table:veri_match}, directly applying our method for single-step fingerprint registration achieves state-of-the-art matching performance than previous methods on FVC2004 datasets. By further using our method as a replacement for initial registration followed by PDRNet for dense registration, we can achieve even better matching performance.

As shown in Table \ref{table:corr_match} and \ref{table:veri_match}, our method shows particularly notable advantages on FVC2004 DB1\_A, which contains many fingerprints with large-scale distortion deformations, demonstrating our method's superiority in handling fingerprint distortions. For FVC2004 DB2\_A, which contains many wet and dry fingerprints, previous fingerprint registration methods rely on fingerprint preprocessing methods that often perform poorly on these low-quality fingerprints. Our method directly inputs the original fingerprint images, thus avoiding the drawbacks of traditional preprocessing methods to some extent, breaking through previous limitations and achieving certain improvements. As for FVC2004 DB3\_A, which uses thermal sweeping sensors for fingerprint collection - a type of fingerprint data not present in our training set - our method can still achieve leading matching performance, especially showing our method's strong generalization ability.

\begin{figure*}
\centering
\subfloat[FVC2004 DB1\_A] 
{\includegraphics[width=0.33\linewidth]{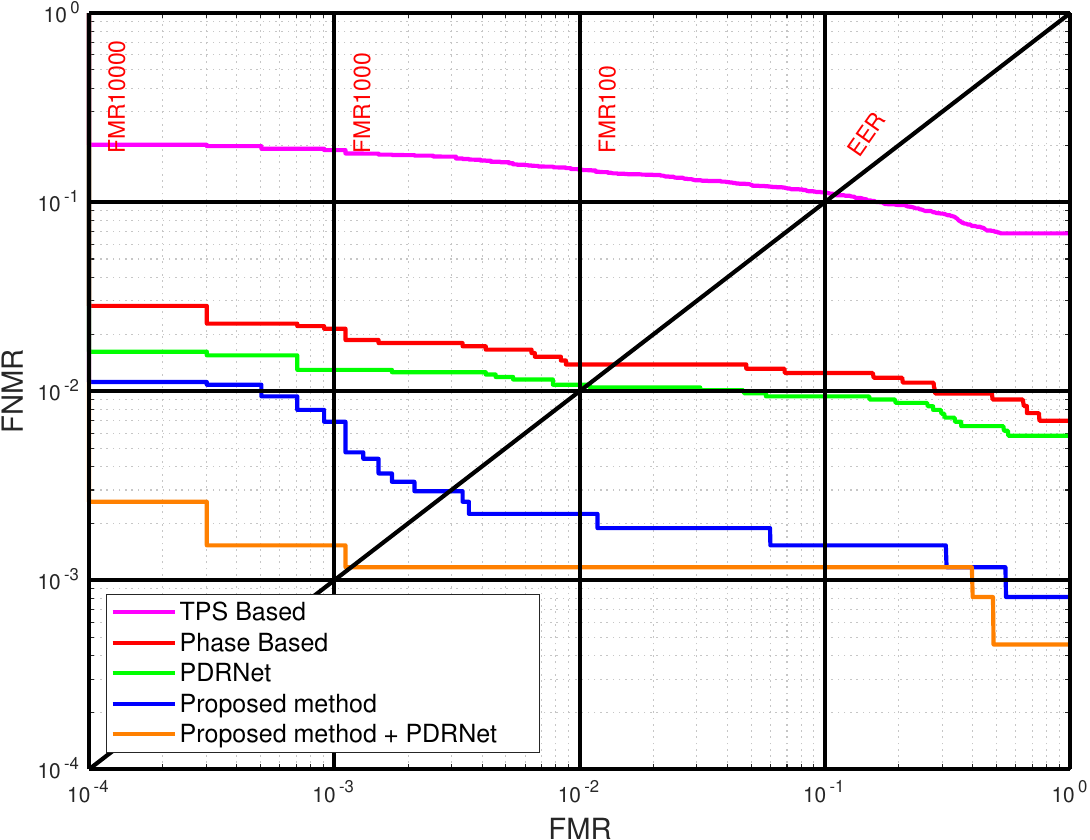}} 
\subfloat[FVC2004 DB2\_A]{\includegraphics[width=0.33\linewidth]{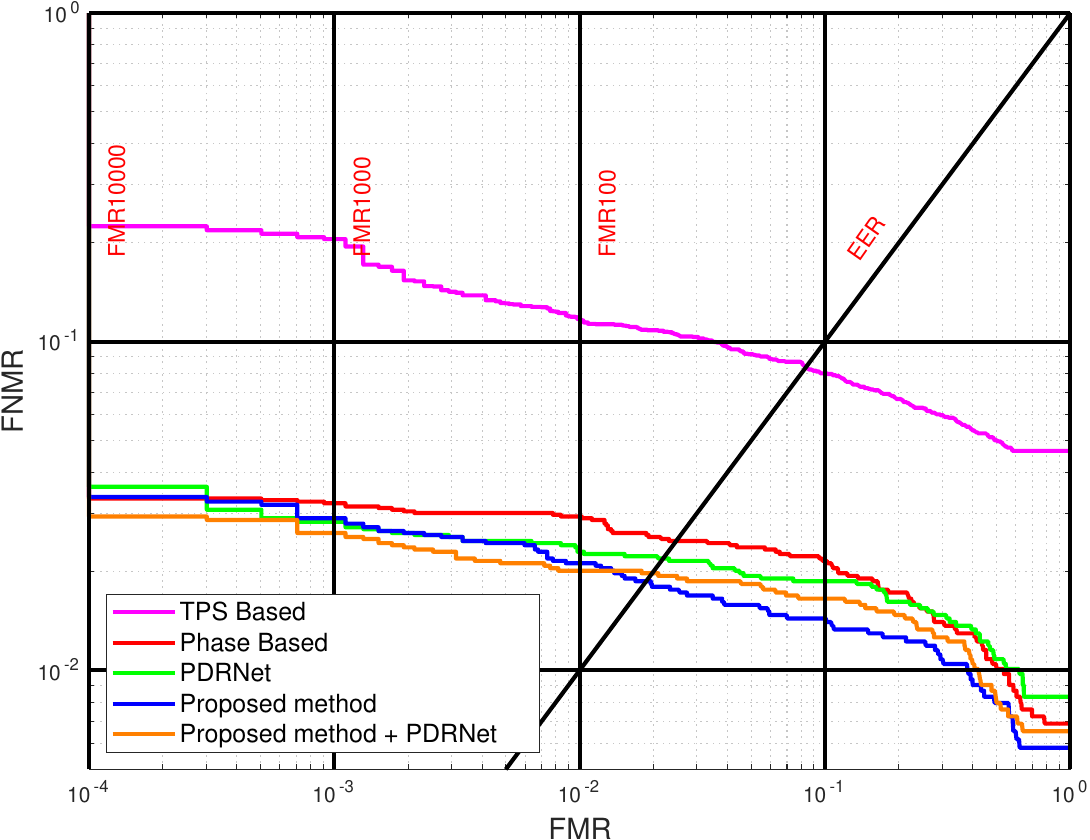}}
\subfloat[FVC2004 DB3\_A]{\includegraphics[width=0.33\linewidth]{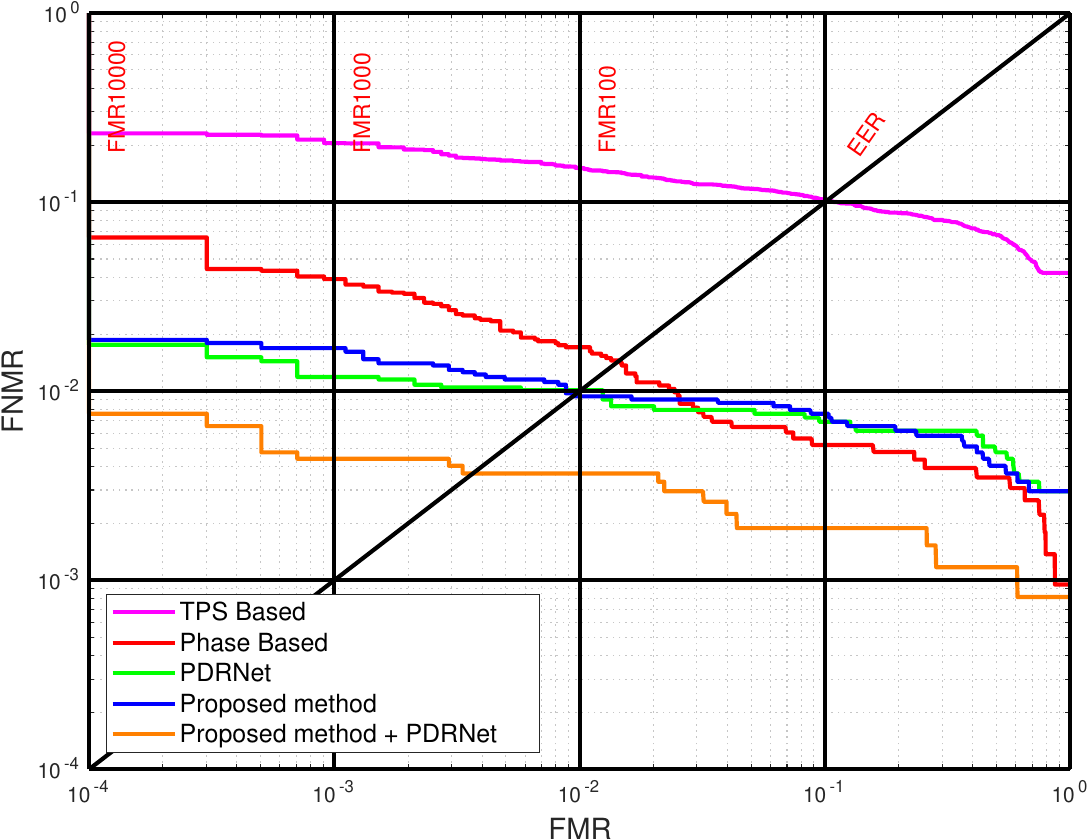}}
\caption{DET curves by image correlator scores on FVC2004.} 
\label{fig:match_corr}  
\end{figure*}

\begin{figure*}
\centering
\subfloat[FVC2004 DB1\_A] 
{\includegraphics[width=0.33\linewidth]{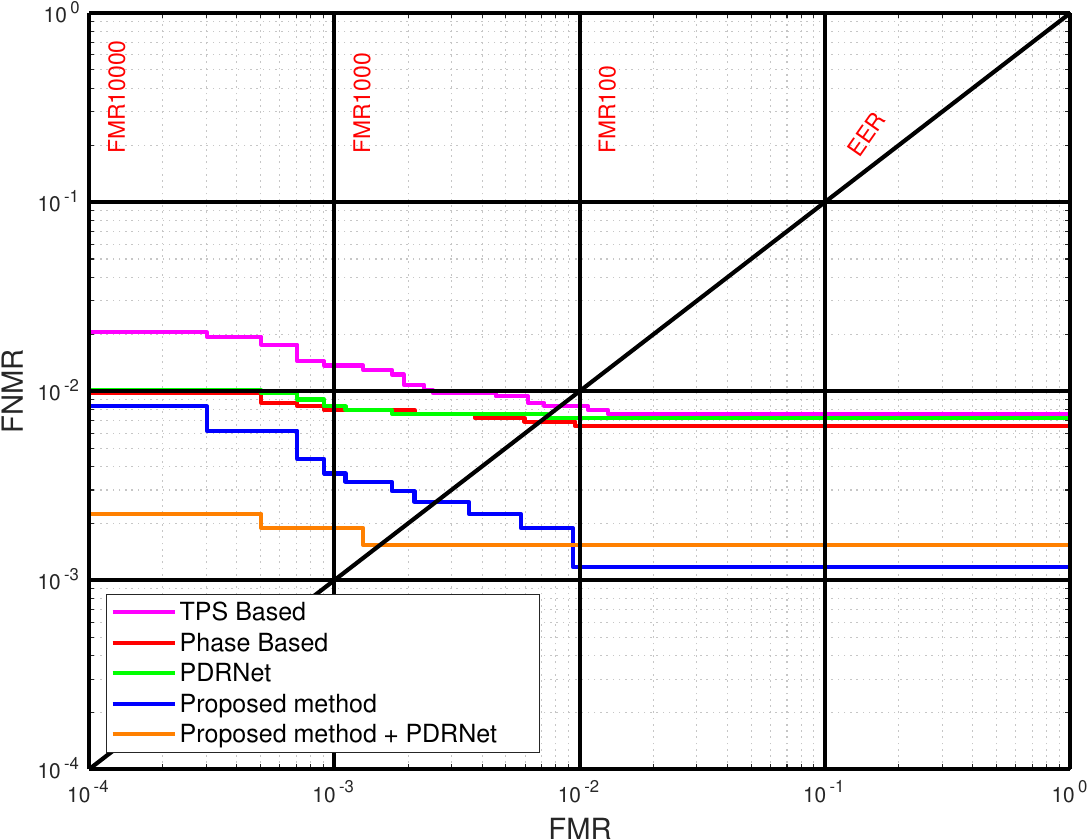}} 
\subfloat[FVC2004 DB2\_A]{\includegraphics[width=0.33\linewidth]{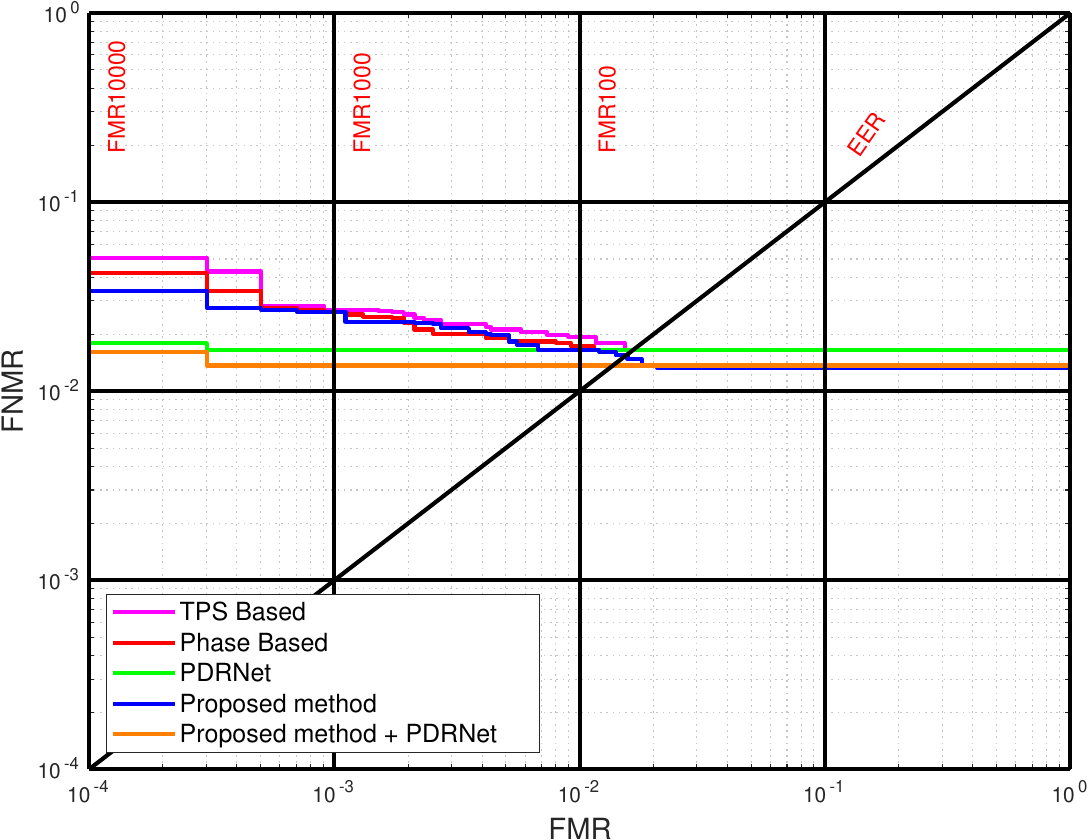}}
\subfloat[FVC2004 DB3\_A]{\includegraphics[width=0.33\linewidth]{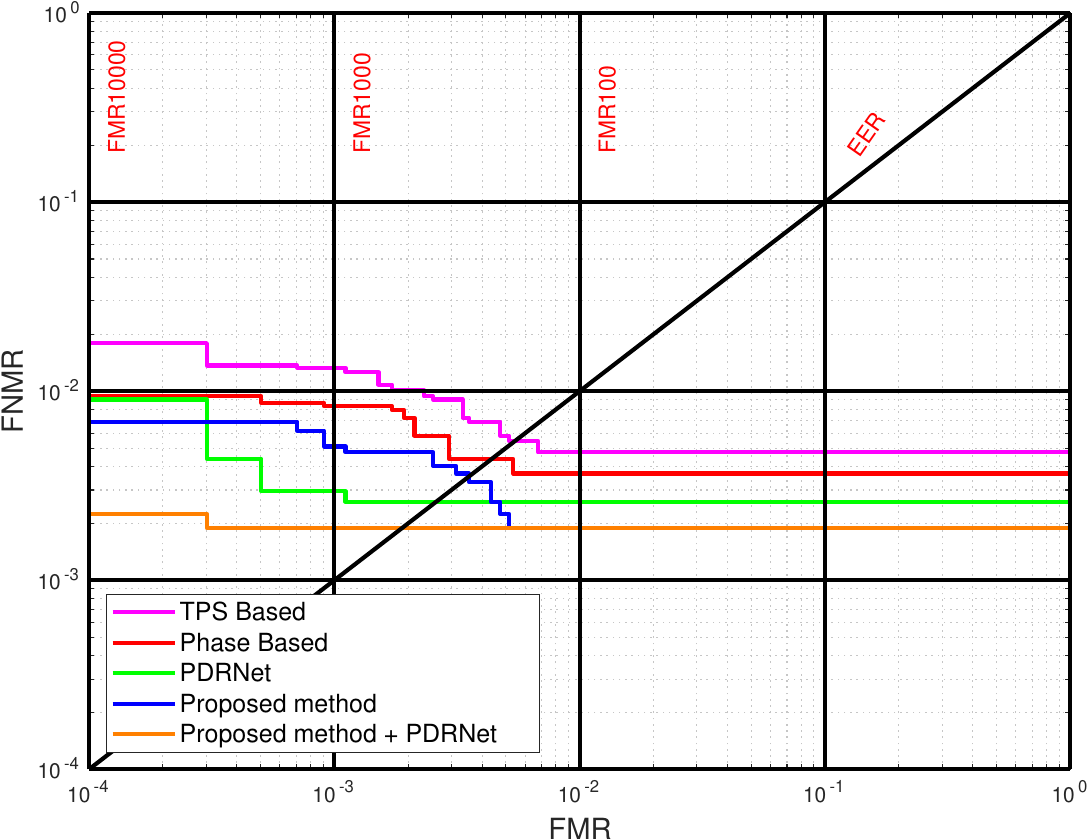}}
\caption{DET curves by Verifinger matcher scores on FVC2004.} 
\label{fig:match_veri}  
\end{figure*}

Fig. \ref{fig:match_corr} and Fig. \ref{fig:match_veri} show the DET curves based on Image Correlator and Verifinger Matcher respectively, providing a more comprehensive view of our experimental results. These also demonstrate that our method has advantages in both single-step and two-step scenarios, particularly showing comprehensive superiority over previous fingerprint registration methods.

We also test our method's performance on latent fingerprints using NIST SD 27 in Table \ref{table:nist27}. It should be noted that our method are not trained on latent fingerprints, while previous deep learning-based fingerprint registration methods are often trained on specific latent fingerprint dataset. Nevertheless, when using Image Correlator, our method still achieves advantages over previous methods in Rank-1 identification rate. As shown in Table \ref{table:nist27}, both in single stage and two stage, our method achieves a Rank-1 identification rate higher than previous methods, also showing our method's advantages on latent fingerprints.

For the recently proposed Transformer-based fingerprint matching method \cite{qiu2024ifvit}, we also compared our method's matching performance with theirs. Given that their method focuses on fixed-length fingerprint representation while our method targets pixel-level alignment, the neural networks are designed and trained with different objectives, making it difficult to conduct fair comparisons using the same matching scores. Since fingerprint matching tasks typically focus primarily on final matching accuracy, we directly compare Image Correlator with their fixed-length representation similarity. As shown in Table \ref{table:ifvit}, our method's EER is significantly lower than IFViT on FVC2004 DB1\_A and DB3\_A, while slightly higher on FVC2004 DB2\_A. This is mainly because our method primarily targets finger deformation, while IFViT incorporated data augmentation for dry and wet fingerprints during training for fingerprint representation. Overall, our method demonstrates advantages in fingerprint matching compared to theirs. Notably, our method use far less training data than theirs - their training data totaled 3,674,027 pairs, while we only used 46,974 pairs.
\begin{table}[]
\centering

\caption{Matching Performance by \textbf{Image Correlator} with Different Fingerprint Registration Algorithms}
\label{table:corr_match}

\begin{tabular}{lcccccc}
\toprule

\multicolumn{1}{l}{}                  & \multicolumn{6}{c}{\textbf{FVC2004}}                                                           \\ \cmidrule(lr){2-7} 
\multicolumn{1}{l}{\textbf{Method}}                                         & \multicolumn{2}{c}{\textbf{DB1\_A}}     & \multicolumn{2}{c}{\textbf{DB2\_A}} & \multicolumn{2}{c}{\textbf{DB3\_A}} \\ \cmidrule(lr){2-3} \cmidrule(lr){4-5} \cmidrule(lr){6-7} 
\multicolumn{1}{l}{}                                         & FMR           & EER           & FMR          & EER        & FMR         & EER         \\ \midrule

TPS Based               & 20.07           & 10.93           & 22.43            & 8.36           & 23.07           & 10.18           \\
Phase Based\cite{cui20182}             & 2.61            & 1.36            & 3.32             & 2.46           & 6.49            & 1.44            \\
PDRNet\cite{guan2024phase}                  & 1.61            & 1.07            & 3.61             & 2.18           & 1.75            & 1.00            \\ \midrule
Proposed                & 1.11   & 0.29   & 3.36             & \textbf{1.86}           & 1.86            & 0.93            \\ \midrule
Proposed + PDRNet     & \textbf{0.25}   & \textbf{0.11}   & \textbf{2.93}             & 1.96           & \textbf{0.86}            & \textbf{0.36}       
\\\bottomrule
\end{tabular}
\begin{tablenotes}
\item FMR represents ZeroFMR.
\end{tablenotes}
\end{table}

\begin{table}[]
\caption{Matching Performance by \textbf{Verifinger Matcher} with Different Fingerprint Registration Algorithms}
\label{table:veri_match}
\centering

\begin{tabular}{lcccccc}
\toprule

\multicolumn{1}{l}{}                  & \multicolumn{6}{c}{\textbf{FVC2004}}                                                           \\ \cmidrule(lr){2-7} 
\multicolumn{1}{l}{\textbf{Method}}                                         & \multicolumn{2}{c}{\textbf{DB1\_A}}     & \multicolumn{2}{c}{\textbf{DB2\_A}} & \multicolumn{2}{c}{\textbf{DB3\_A}} \\ \cmidrule(lr){2-3} \cmidrule(lr){4-5} \cmidrule(lr){6-7} 
\multicolumn{1}{l}{}                                         & FMR           & EER           & FMR          & EER        & FMR         & EER         \\ \midrule

TPS Based               & 2.04           & 0.82          & 5.07            & 1.64           & 1.79           & 0.54           \\
Phase Based\cite{cui20182}             & 0.96           & 0.68          & 4.21             & 1.64           & 0.93            & 0.43            \\
PDRNet\cite{guan2024phase}  & 1.00           & 0.73          & 1.79             & 1.64           & 0.89            & 0.25            \\ \midrule
Proposed                & 0.82           & 0.25          & 3.39             & 1.54           & 0.68            & 0.34            \\ \midrule
Proposed + PDRNet       &\textbf{0.21}   &\textbf{0.14}  &\textbf{1.61}     & \textbf{1.36}  & \textbf{0.21}   & \textbf{0.18}       
\\\bottomrule
\end{tabular}
\begin{tablenotes}
\item FMR represents ZeroFMR.
\end{tablenotes}
\end{table}


\begin{table}[]
\centering

\caption{Rank-1 Identification Rate by Image Correlator with Different Fingerprint Registration Algorithms on NIST SD 27}
\label{table:nist27}
\begin{tabular}{lc}
\toprule
\textbf{Method}             & \textbf{Rank-1} \\ \midrule
TPS Based          & 0.02   \\
Phase Based & 0.39   \\ 
PDRNet & 0.44   \\  \midrule
Proposed           & \textbf{0.45}  \\ \midrule
Proposed + PDRNet  & \textbf{0.53}   \\ \midrule
\end{tabular}
\end{table}

\begin{table}[]
\centering

\caption{EER vs Interpretable Fingerprint Matching\cite{qiu2024ifvit} }
\label{table:ifvit}

\begin{tabular}{lccc}
\toprule

\multicolumn{1}{l}{}                  & \multicolumn{3}{c}{\textbf{FVC2004}}                                                           \\ \cmidrule(lr){2-4} 
\multicolumn{1}{l}{\textbf{Method}}                                         & \multicolumn{1}{c}{\textbf{DB1\_A}}     & \multicolumn{1}{c}{\textbf{DB2\_A}} & \multicolumn{1}{c}{\textbf{DB3\_A}} \\ \cmidrule(lr){2-2} \cmidrule(lr){3-3} \cmidrule(lr){4-4} 
\multicolumn{1}{l}{}                                                    & EER                     & EER                 & EER         \\ \midrule

IFViT \cite{qiu2024ifvit}              & 1.87           & \textbf{1.83}           & 1.49                       \\
Proposed                & \textbf{0.29}   & 1.86   & \textbf{0.93}                       
\\\bottomrule
\end{tabular}
\end{table}

\subsection{Registration Accuracy}
Our improvements mainly focus on addressing registration failure issues in previous initial registration methods that led to subsequent dense registration. Therefore, when performing our single-step registration, although we can achieve good Matching Performance, there still exists small gaps in registration accuracy compared to those two-step dense registration methods. In the two-step scenario where our method serves as initial registration, as shown in Table \ref{table:reg_acc}, our method still achieves the best Verifinger Score across all three datasets, and the best Image Correlator score on FVC2004 DB1\_A, with close-to-best results on the other two datasets. 

\begin{figure*}[!]
\centering
\centerline{\includegraphics[width=\linewidth]{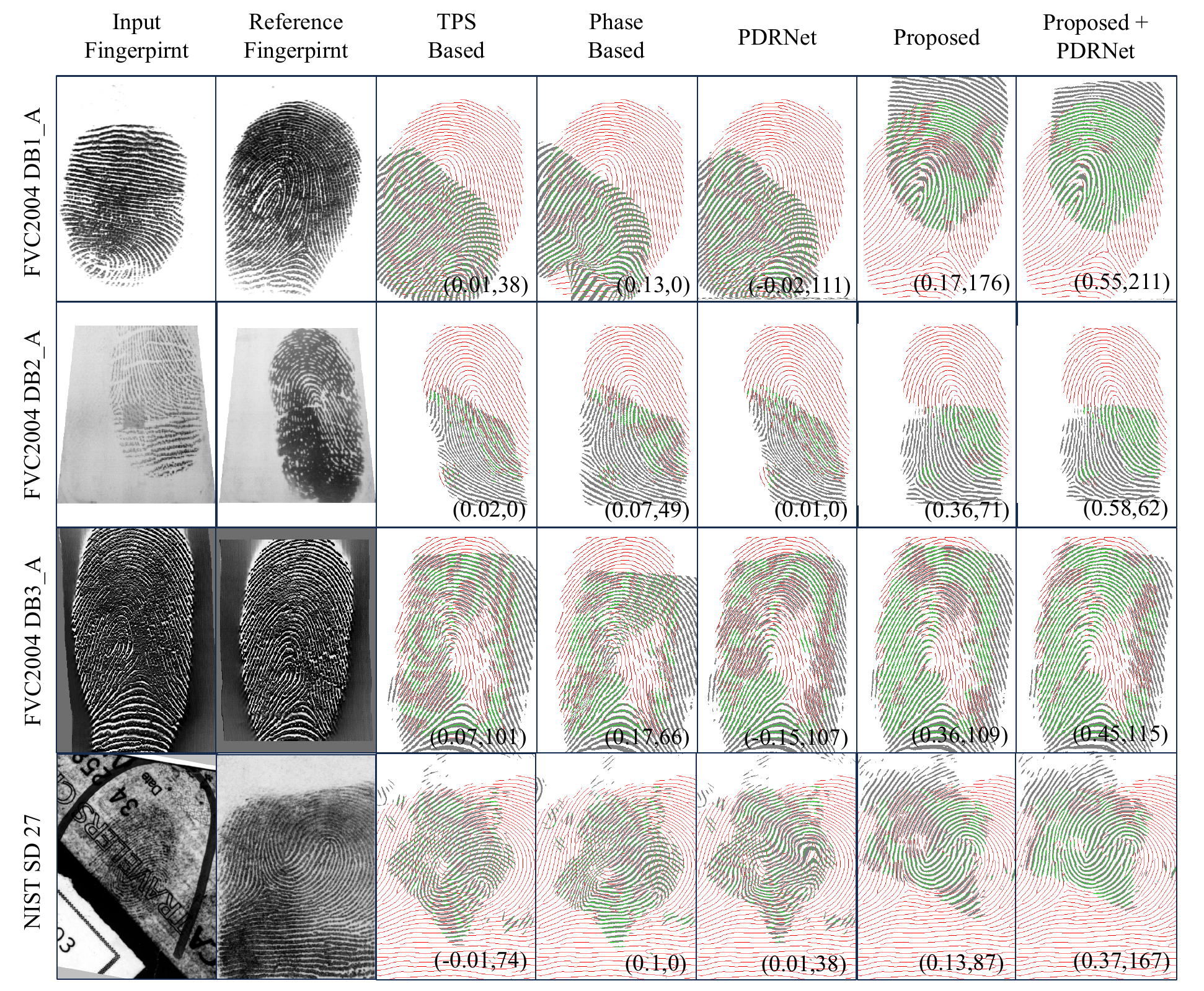}}
\caption{Examples of fingerprint registration for genuine matching fingerprints. The beginning of each row gives the name of corresponding datasets, and numbers in brackets are matching scores by image correlator and VeriFinger. Green indicates overlap, while red and gray indicate non-overlapping areas of respective fingerprints. The figure shows that our method can handle data that previous fingerprint registration methods are unable to align.}
\label{fig:match_example}
\end{figure*}

\begin{table}[]

\caption{Registration Accuracy with Different Fingerprint Registration Algorithms}
\label{table:reg_acc}

\begin{tabular}{lcccccc}
\toprule

\multicolumn{1}{l}{}                  & \multicolumn{6}{c}{\textbf{FVC2004}}                                                           \\ \cmidrule(lr){2-7} 
\multicolumn{1}{l}{\textbf{Method}}                                         & \multicolumn{2}{c}{\textbf{DB1\_A}}     & \multicolumn{2}{c}{\textbf{DB2\_A}} & \multicolumn{2}{c}{\textbf{DB3\_A}} \\ \cmidrule(lr){2-3} \cmidrule(lr){4-5} \cmidrule(lr){6-7} 
\multicolumn{1}{l}{}                                         & NCC           & VF           & NCC          & VF        & NCC         & VF         \\ \midrule

TPS Based               & 0.28           & 249          & 0.29            & 224           & 0.23           & 231           \\
Phase Based\cite{cui20182}             & 0.72            & 271           & \textbf{0.68}             & 242           & \textbf{0.69}            & 253            \\
PDRNet\cite{guan2024phase}                  & 0.68            & 273            & 0.62             & 241           & 0.60            & 258            \\ \midrule
Proposed                & 0.55           & 267              & 0.50             & 230           & 0.45            & 249            \\ \midrule
Proposed + PDRNet     & \textbf{0.73}   & \textbf{279}   & 0.67             & \textbf{243}           & 0.66            & \textbf{261}       
\\\bottomrule
\end{tabular}
\end{table}
Fig. \ref{fig:match_example} shows the registration results of some true matching fingerprints on several fingerprint datasets. Due to the limitations of the initial registration algorithm, both traditional and deep neural network-based fingerprint dense registration algorithms were unable to handle images with multiple quality issues. The first row shows the registration of fingerprints with large distortion and small overlap area, the second row shows the registration of wet fingerprints with small overlap area, the third row shows the registration of a pair of low-quality fingerprints, and the fourth row shows examples where our method outperforms previous methods on latent fingerprints registration. Our method achieves the best registration performances on low quality fingerprints that previous initial registration method can not handle.

\subsection{Ablation Study}

We conduct ablation experiments on FVC2004 DB1\_A, as shown in Table \ref{table:ablation}, using Image Correlator's ZeroFMR as the metric. The ablation experiments covered four aspects: 

1) Fine-Level Matching: We remove the network's fine level rectification component (introduced in Section \ref{sec:Fine-Level}), and results in a significant decrease in network matching performance, indicating that coarse level feature matching alone is insufficient for pixel-level registration. 

2) Self and Cross Attention in matching network architecture\cite{sun2021loftr}\cite{qiu2024ifvit}: We also train a registration network under similar architecture, which only lacks local information processing. As a result, this network struggles to achieve state-of-the-art matching performance. We first test the altered registration network with the same test hyperparameters, and its FMR is even worse than the TPS Based method. We manage to adjust the hyperparameters to achieve the best Matching Performance, but its performance still shows a significant gap compared to the original proposed registration network. 

3) Window size: We find that using the same fine registration window size $w$ during testing as in training make it difficult to obtain correct registration results. We increase $w$ at $w=17$, that can already achieve a lower FMR. When $w =25$, the matching performance on FVC2004 DB1\_A reaches state-of-the-art levels. Further increasing $w$ continue to decrease FMR, but EER also increase, indicating the emergence of some false matching points that would affect further dense registration, so we ultimately chose $w =25$. 

4) The TPS regularization term $\lambda$: We first prove that without regularization, i.e., when $\lambda=0$, our network struggles to perform well. As we gradually increase $\lambda$, we achieve the best matching performance when its value is 0.2.

\begin{table*}[]

\centering
\caption{Ablation Study of the Proposed Network with Different Modules and Strategies on FVC 2004 DB1\_A}
\label{table:ablation}

\begin{tabular}{cccccccccc}
\toprule
\multicolumn{9}{c}{\textbf{Modules \& Strategies}}                                                                                                                & \multirow{3}{*}{FMR} \\ \cmidrule(lr){1-9}
\multirow{3}{*}{\textbf{Fine-Level Matching}} & \multicolumn{2}{c}{\textbf{Coarse Matching Network}}       & \multicolumn{3}{c}{\textbf{Fine-Window Size $w$}} & \multicolumn{3}{c}{\textbf{TPS Regularization Paramater $\lambda$}} &                      \\ \cmidrule(lr){2-3}\cmidrule(lr){4-6}
\cmidrule(lr){7-9}
                    & Self and Cross Attention & Global-Local Attention & 11         & 17         & 25         & 0          & 0.2         & 0.4         &                      \\ \midrule
              -      &    -                      & \checkmark                      &    -        &    -        & -{$\dag$}          &    -        & \checkmark           &     -        & 5.00                 \\
\checkmark                   & \checkmark                       &    -                    &      -      &       -     & \checkmark          &        -    & \checkmark           &     -        &  24.21                \\
\checkmark                   & \checkmark*                        &   -                     &         -   &       -     & -*          &    -        & \checkmark           &     -        & 3.14             \\
\checkmark                   &     -                     & \checkmark                      & \checkmark          &     -       &      -      &       -     & \checkmark           &        -     & 36.54                \\
\checkmark                   &         -                 & \checkmark                      &      -      & \checkmark          &     -       &      -      & \checkmark           &         -    & 1.71                 \\
\checkmark                   &                -          & \checkmark                      &      -      &        -    & \checkmark          & \checkmark          &     -        &       -      & 4.82                 \\
\checkmark                   &          -                & \checkmark                      &     -       &       -     & \checkmark          &      -      &       -      & \checkmark           & 1.36                 \\
\checkmark                   &               -           & \checkmark                      &        -    &       -     & \checkmark          &     -       & \checkmark           &          -   & \textbf{1.11}                 \\ \bottomrule
\end{tabular}
\begin{tablenotes}
\item {$\dag$}: There is no need to configure the fine-window size when precise matching is not required.
\item *: The performance of the two different coarse matching networks is significantly influenced by hyperparameters. The parameters for other ablation experiments are controlled, but to achieve better results with the coarse matching network consisting of Self and Cross Attention, this set of parameters differs considerably from the others.
\end{tablenotes}
\end{table*}

\subsection{Efficiency}

We test the running time of four methods on FVC2004 DB1\_A using the same server, measuring the time taken to complete image deformation. For the previous dense registration methods, only the second fine registration step time is recorded. The image size is $640\times 480$. Testing is conducted on a 2.5 GHz CPU and an NVIDIA GeForce RTX 4090 GPU. Our method is significantly more efficient than previous dense registration algorithms as shown in Table \ref{table:efficiency}.
\begin{table}[]
\centering

\caption{Time Cost of Different Fingerprint Registration Algorithms for Processing a 640 × 480 Fingerprint Pair in FVC2004 DB1\_A}
\label{table:efficiency}
\begin{tabular}{lc}
\toprule
\textbf{Method}             & \textbf{Time (s)} \\ \midrule
TPS Based          & 0.33\\

Phase Based           & 20.24 \\
PDRNet & 1.04\\ \midrule
Proposed & \textbf{0.19}   \\
\bottomrule
\end{tabular}
\end{table}
\section{CONCLUSION}
This paper proposes a Transformer-based single-step fingerprint registration method for accurate and efficient fingerprint alignment. The proposed method adopts a Global-Local Attention Module, which improves both the registration and matching performances than previous two-step registration methods. The proposed method can successfully registers fingerprint pairs that could not be registered in previous registrations due to quality issues, overcoming the bottleneck of traditional fingerprint registration algorithms. However, our method still faces challenges in high-level accuracy as a single-step approach. In the future, we plan to further integrate the fingerprint prior knowledge used in dense registration algorithms and attempt to design a more powerful single-step fingerprint registration algorithm.
\section*{Acknowledgments}

We would like to appreciate Mr. Xiongjun Guan for providing
us with the code of PDRNet for experiments.
This work is supported in part by the National Natural
Science Foundation of China under Grants 62206026.







 

\bibliographystyle{IEEEtran}
\bibliography{ref}









\end{document}